\documentclass[final]{cvpr}

\usepackage{times}
\usepackage{epsfig}
\usepackage{graphicx}
\usepackage{amsmath}
\usepackage{amssymb}
\usepackage{csquotes}
\usepackage{bbding}

\usepackage[normalem]{ulem}

\usepackage[font=small,labelfont=bf,tableposition=top]{caption}

\usepackage[pagebackref=true,breaklinks=true,colorlinks,bookmarks=false]{hyperref}

\def\ourmodel{Rank-Net}

\def\cvprPaperID{765} 
\def\confYear{CVPR 2021}

\begin{document}

\title{Simultaneously Localize, Segment and Rank the Camouflaged Objects}

\author{
Yunqiu Lv$^{1,\ddag}$~
Jing Zhang$^{2,3,\ddag}$~
Yuchao Dai$^1$\href{mailto:daiyuchao@gmail.com}{\Envelope}~
Aixuan Li$^{1}$~
Bowen Liu$^{1}$~ 
Nick Barnes$^2$~
Deng-Ping Fan$^4$\\
$^1$ Northwestern Polytechnical University, China \quad
$^2$ Australian National University, Australia\\
$^3$ CSIRO, Australia \quad
$^4$ Inception Institute of AI (IIAI), Abu Dhabi, UAE\\
{\tt \small $\ddag$ Equal contributions; \Envelope~Corresponding author:daiyuchao@nwpu.edu.cn
}
\\
}

\twocolumn[{%
\renewcommand\twocolumn[1][]{#1}%
\maketitle
\begin{center}
    \centering
   \begin{center}
   \vspace{-5mm}
   \begin{tabular}{{c@{ } c@{ } c@{ } c@{ }}}
   {\includegraphics[width=0.21\linewidth, height=0.16\linewidth]{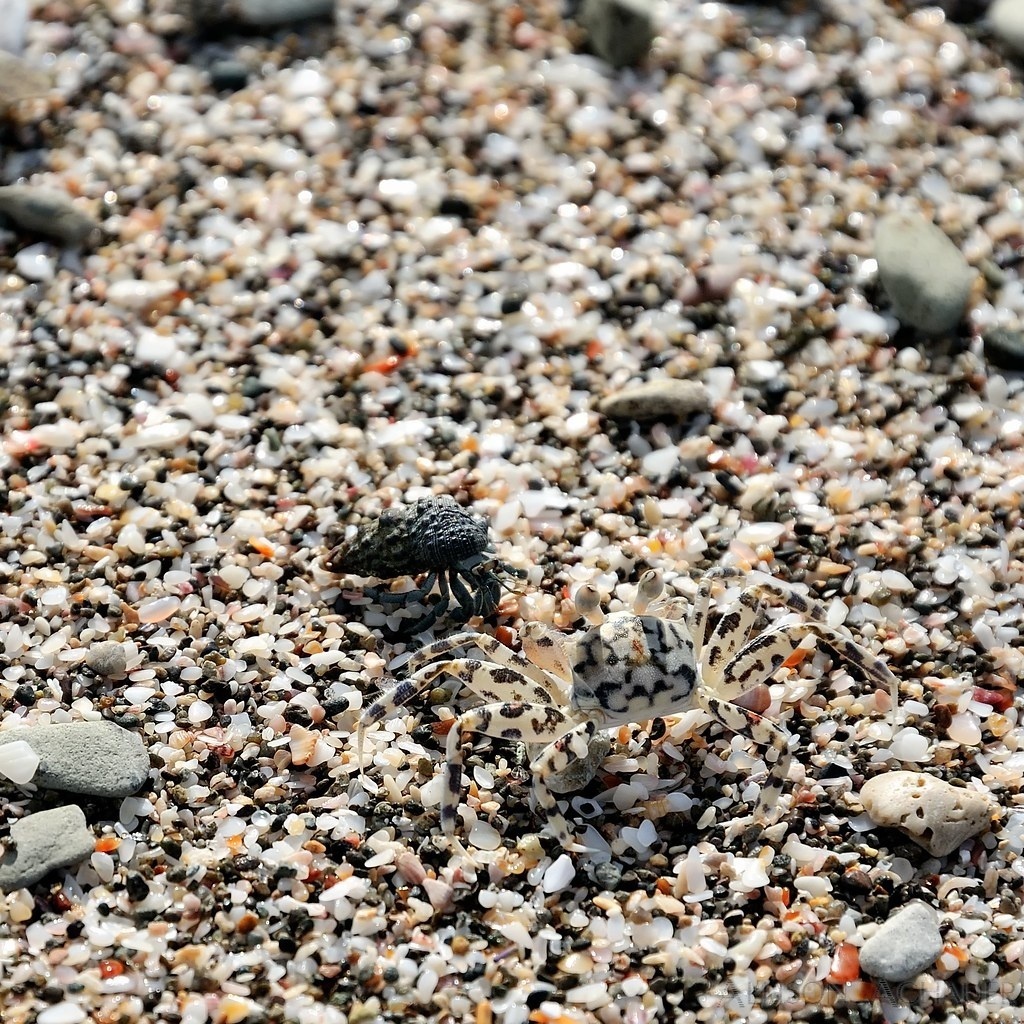}}&
    {\includegraphics[width=0.21\linewidth, height=0.16\linewidth]{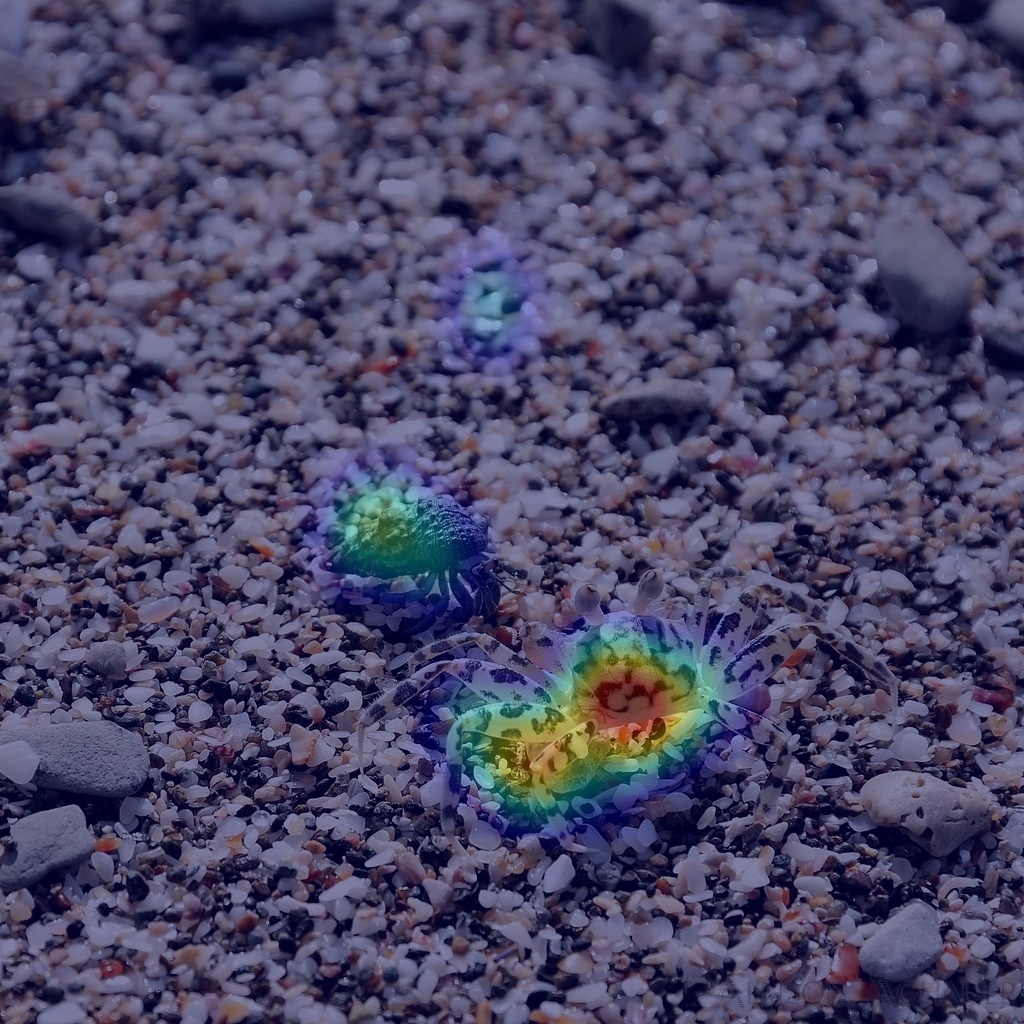}}&
    {\includegraphics[width=0.21\linewidth, height=0.16\linewidth]{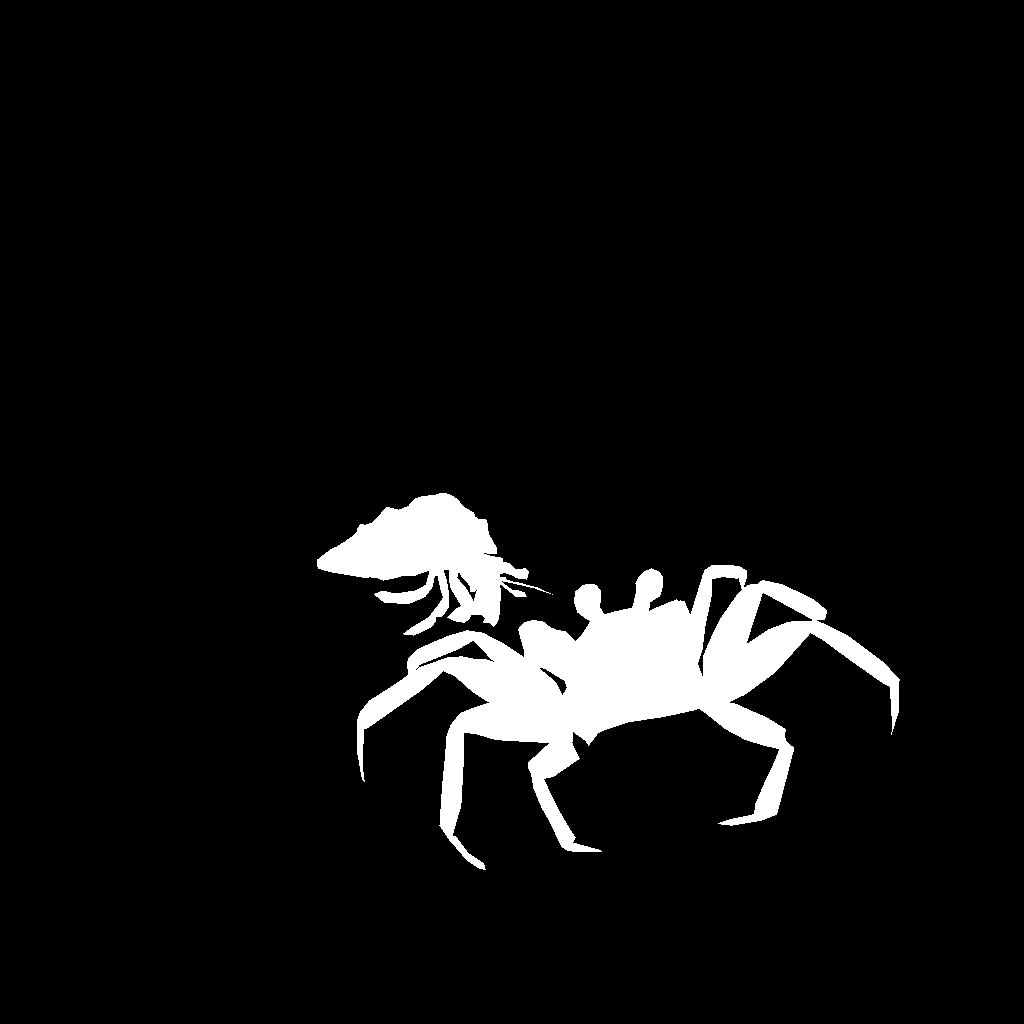}}&
    {\includegraphics[width=0.21\linewidth, height=0.16\linewidth]{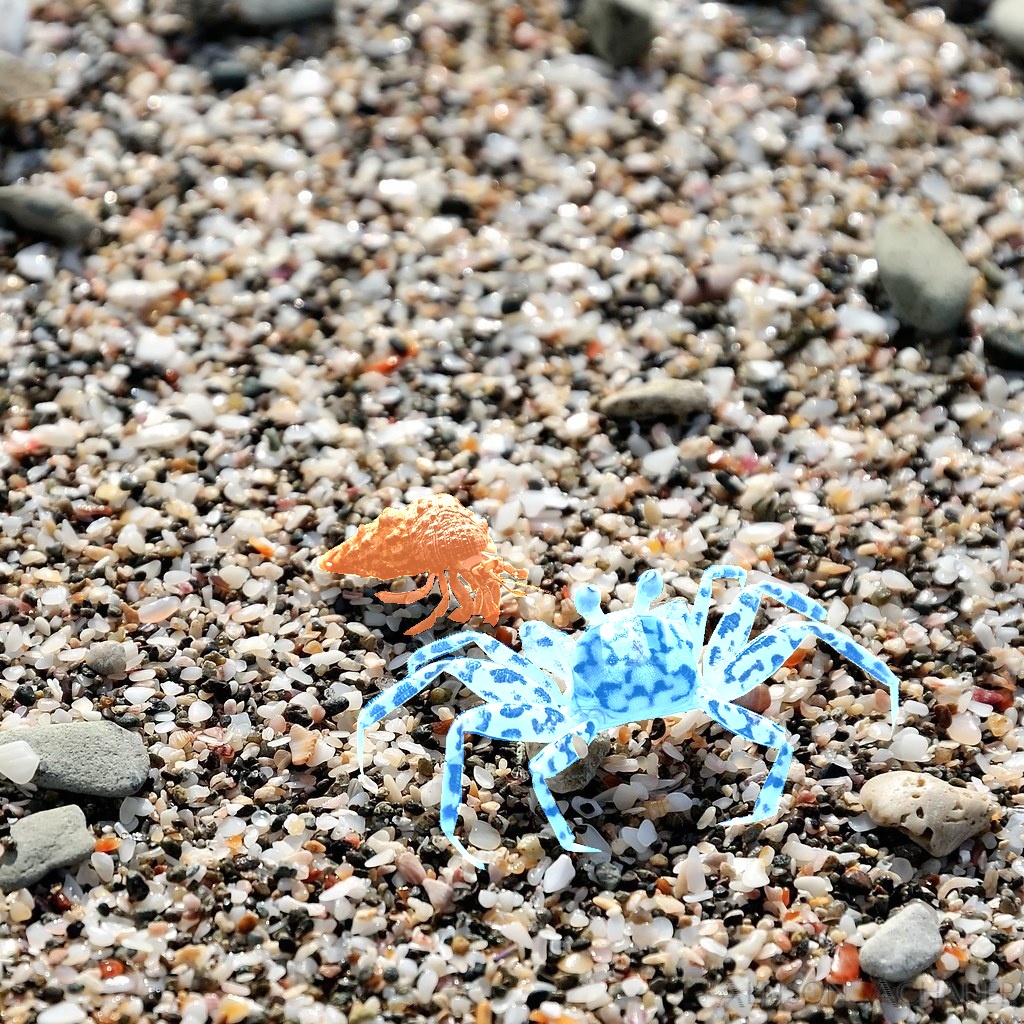}} \\
   {\includegraphics[width=0.21\linewidth, height=0.16\linewidth]{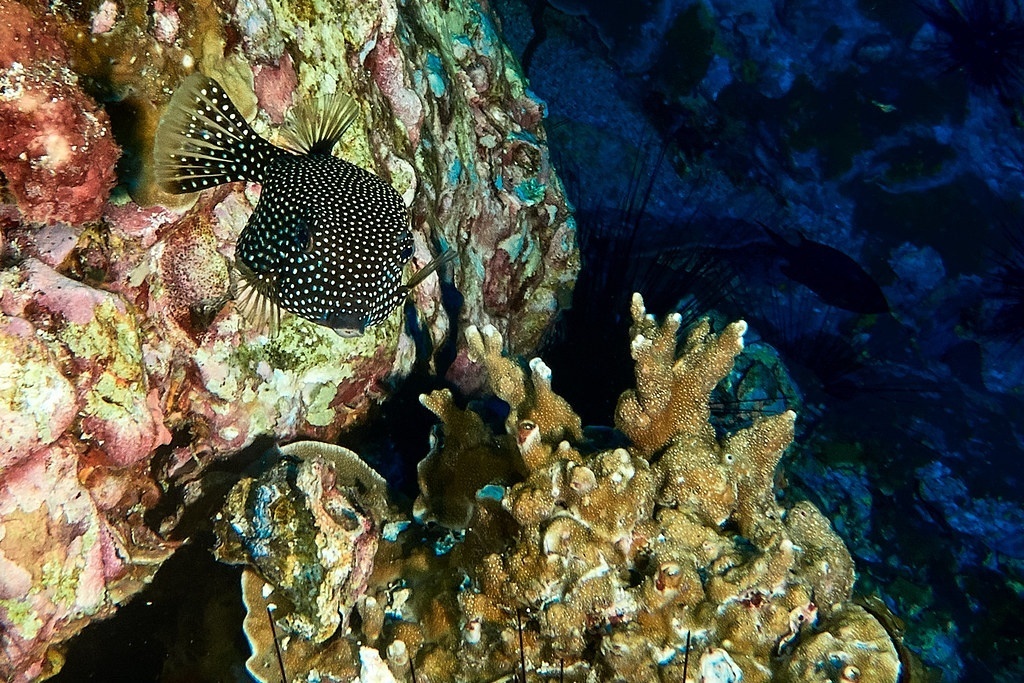}}&    
   {\includegraphics[width=0.21\linewidth, height=0.16\linewidth]{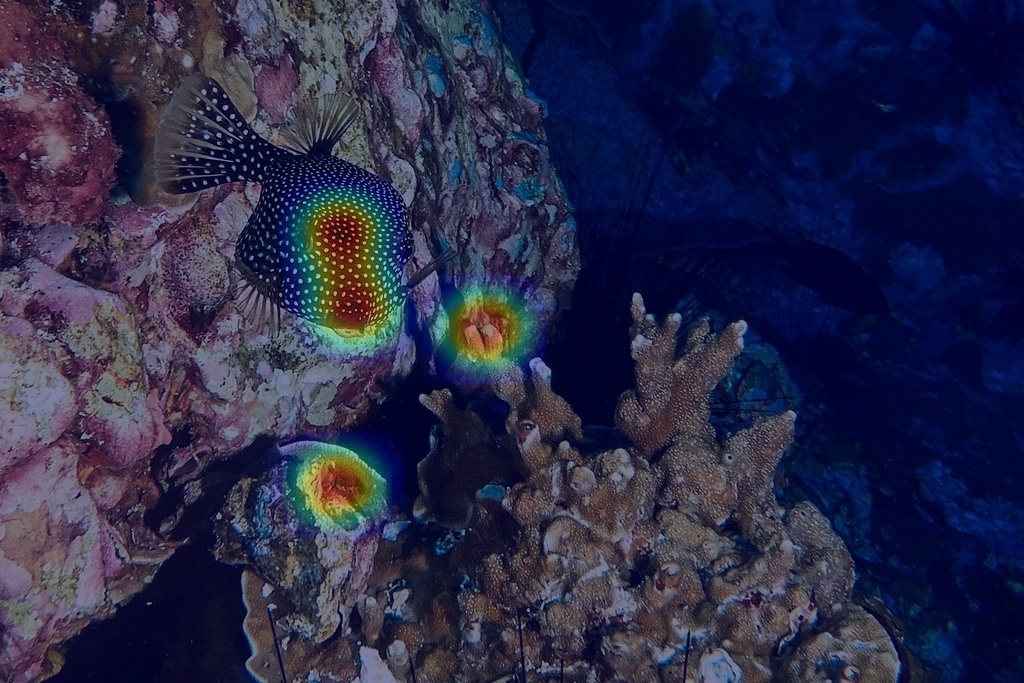}}&
    {\includegraphics[width=0.21\linewidth, height=0.16\linewidth]{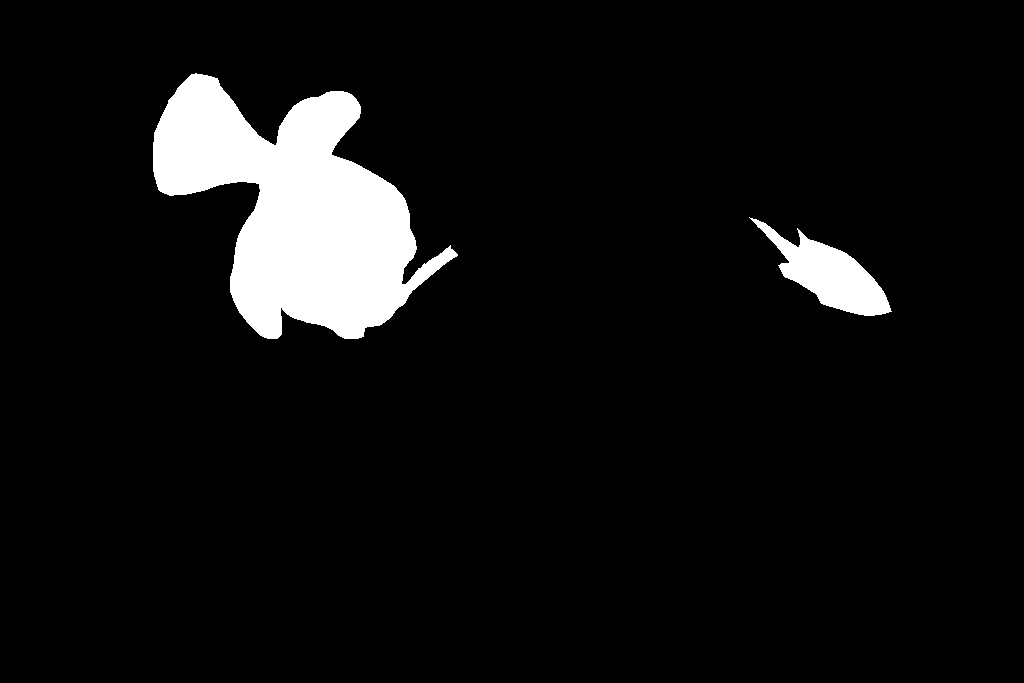}}&
    {\includegraphics[width=0.21\linewidth, height=0.16\linewidth]{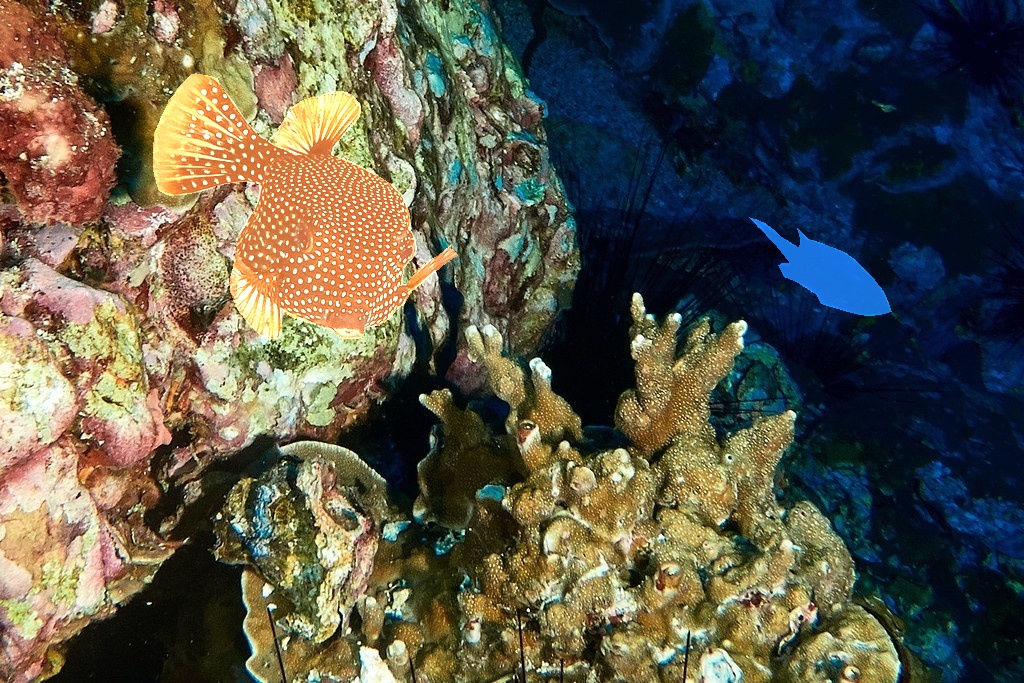}} \\
    \footnotesize{Image} &
    \footnotesize{Fixation} & \footnotesize{Binary} & \footnotesize{Ranking} \\
   \end{tabular}
   \end{center}
    \captionof{figure}{
    The conventional \enquote{Binary} ground truth only provides the scope of the camouflaged objects. We present additional 
    fixation (\enquote{Fixation}) and ranking (\enquote{Ranking}) annotations,
    where the former discovers regions that make camouflaged objects detectable and the latter highlights the level of camouflage. Blue color in \enquote{Ranking} indicates higher rank (harder) of camouflage.
    } 
    \label{sample_show_front_page}
\end{center}%
}
]

\begin{abstract}
Camouflage is a key defence mechanism across species that is critical to survival. 
Common strategies for
camouflage include background matching, imitating the color and pattern of the environment, and disruptive coloration, disguising body outlines \cite{background_matching}. Camouflaged object detection (COD) aims
to segment 
camouflaged objects hiding in their surroundings. 
Existing COD models are built upon binary ground truth to segment the camouflaged objects without illustrating the level of camouflage.
In this paper, we revisit this task and argue that explicitly modeling the conspicuousness of camouflaged objects against their particular backgrounds can not only lead to a better understanding about camouflage and evolution of animals, but also provide guidance to design more sophisticated camouflage techniques.
Furthermore, we observe that it is some specific parts of the camouflaged objects that make them detectable by
predators. With the above understanding
about camouflaged objects, we present the first ranking based COD network (\ourmodel) to simultaneously localize, segment and rank 
camouflaged objects. The localization model is proposed to find the discriminative regions that make the camouflaged object obvious. The segmentation model segments the full scope of the camouflaged objects. 
Further,
the ranking model infers the detectability of different camouflaged objects.
Moreover, we contribute a large COD testing set to evaluate the generalization ability of COD models.
Experimental results show that our model achieves new state-of-the-art,
leading to a more interpretable COD network\footnote{Our code and data is publicly available at: \url{https://github.com/JingZhang617/COD-Rank-Localize-and-Segment}. More detail about the training dataset can be found in \url{http://dpfan.net/camouflage}.}.

\end{abstract}

\section{Introduction}
Camouflage is one of the most important anti-predator defences that prevents the prey
from being recognized by predators \cite{skelhorn2016cognition}. Two main strategies have been widely used among 
prey
to become camouflaged,
namely background matching and disruptive coloration \cite{background_matching}. The prey
that rely on the former approach usually share similar color or pattern with their
habitats, while for complex habitats, the background matching approach may increase their visibility. 
Disruptive coloration works better in complex environments, where 
prey
evolve to have relative high contrast markings near the body edges.

Both background matching and disruptive coloration aim
to hide 
prey
in the environment, or greatly reduce their saliency,
which is closely related to the perception and cognition of
perceivers. 
By delving into the process of camouflaged object detection, the mechanisms of the human visual system can
be finely explored. Further, an effective camouflaged object detection model has potential to be applied in the field of agriculture for insect control, or in medical image segmentation to detect an
infection or tumor area~\cite{fan2020pranet,fan2020inf}. Further, it can improve performance for general object detection, for example where objects appear against similar backgrounds \cite{fan2020camouflaged}.




Existing camouflaged object detection models \cite{fan2020camouflaged,le2019anabranch, zhai2021Mutual, mei2021Ming, aixuan2021joint} are designed based on 
binary ground truth camouflaged object datasets \cite{le2019anabranch,fan2020camouflaged,Chameleon2018} as shown in Fig.~\ref{sample_show_front_page}, which can only reveal the existence of the camouflaged objects without illustrating the level of camouflage. We argue that the estimation of the conspicuousness of camouflaged object against its surrounding can lead to a better understanding about evolution of animals. Further, understanding the level of camouflage can help to design more sophisticated camouflage techniques
\cite{background_matching}, thus 
the prey
can avoid being detected by 
predators.
To model the detectability of camouflaged objects, we introduce the first camouflage ranking model to infer the level of camouflage. Different from existing binary ground truth based models \cite{fan2020camouflaged,le2019anabranch, zhang2020learning},
we can produce the instance-level ranking-based camouflaged object prediction, indicating the global difficulty for human to observe the camouflaged objects.


Moreover, since most 
camouflaged objects lack obvious contrast with the background in terms of low-level features \cite{stevens2009animal}, the detection of camouflaged objects may resort to features relevant to some \enquote{discriminative patterns}, such as face, eyes or antenna. We argue that it is those \enquote{discriminative patterns} that make the
prey
apparent to 
predators. For background matching, these patterns 
have different colors to
the surroundings, and for disruption coloration, they are low contrast body outlines in the complex habitats.
To better understand the camouflage attribute of prey,
we also propose to reveal the most detectable region of the camouflaged objects, namely the camouflaged object discriminative region localization.

As there exists no ranking based camouflaged object detection dataset, we relabel an existing camouflaged object dataset \cite{fan2020camouflaged,le2019anabranch} with an eye tracker to record the detection delay\footnote{We define the median time for multiple observers to notice each camouflaged instance as the detection delay for this instance.} 
of each camouflaged instance. We assume that the longer it takes for the observer to notice the camouflaged object, the higher level of this camouflaged instance. Taking a fixation based camouflaged object detection dataset, we obtain the ranking dataset based on the detection delay, as shown in Fig.~\ref{sample_show_front_page}.
At the same time, the fixation dataset can be used to estimate the discriminative regions of the camouflaged objects.

As far as we know, there only exists one large camouflaged object testing dataset, the COD10K \cite{fan2020camouflaged}, while the sizes of other testing datasets \cite{le2019anabranch,Chameleon2018} are less than 300. We then contribute another camouflaged object testing dataset, namely NC4K, which includes 4,121 images downloaded from the Internet.
The new testing dataset
can be used to evaluate the generalization ability of existing models.

Our main contributions can be summarized as: 1) We introduce 
camouflaged object ranking (COR) and camouflaged object localization (COL) as two new tasks to estimate the difficulty of camouflaged objects and identify the regions that make the camouflaged object obvious. 2) We provide corresponding training and testing datasets for the above two tasks. We also contribute the largest camouflaged object detection testing dataset. 3) We propose a triplet tasks learning model to simultaneously localize, segment and rank the camouflaged objects.
\section{Related Work}

\noindent\textbf{Camouflaged object detection dataset:}
There mainly exist
three camouflaged object detection datasets, namely the CAMO \cite{le2019anabranch} dataset, the CHAMELEMON \cite{Chameleon2018} dataset and the COD10K \cite{fan2021concealed,fan2020camouflaged} dataset. The CAMO dataset \cite{le2019anabranch} includes 1,250 camouflaged images divided into eight categories, where 1,000 camouflaged images are for training, and the remaining 250 images are for testing. 
The CHAMELEON dataset \cite{Chameleon2018} has 76 images downloaded from the Internet for testing.
Fan \etal \cite{fan2020camouflaged} provided a more challenging dataset, named COD10K.
They released 3,040 camouflaged images for training and 2,026 images for testing. Compared with existing camouflaged object datasets, which include only the binary ground truth, we provide extra ranking-based and discriminative region-based annotations. Further, we provide the largest testing dataset with 4,121 images for effective model evaluation.

\begin{figure*}[!htp]
   \begin{center}
   \begin{tabular}{c@{ }}
   {\includegraphics[width=0.85\linewidth]{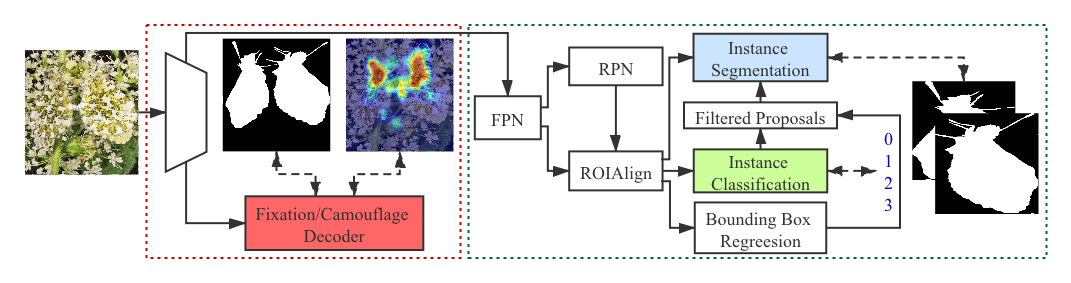}}\\
   \end{tabular}
   \end{center}
  \vspace{-7mm}
   \caption{Overview of the proposed network. We have two main tasks in our framework, namely the camouflaged object ranking which is supervised by the ranking ground truth and each rank based binary segmentation map, and a joint learning framework for camouflaged object discriminative region localization and segmentation. With the input image, our model is trained end-to-end to produce discriminative region localization, camouflaged object segmentation and camouflage ranking. \enquote{FPN} and \enquote{RPN} are the Feature Pyramid Network \cite{fpn} and the Region Proposal Network \cite{faster_rcnn}, respectively.
   }
  \vspace{-3mm}
\label{fig:network_overview}
\end{figure*}

\noindent\textbf{Camouflaged object detection:}
Camouflage is a useful technique for animals to conceal themselves from visual detection by others \cite{merilaita2017camouflage, troscianko2009camouflage}. 
In early research,
most methods use low-level features, including texture, edge, brightness and color features, to discriminate objects from the background \cite{bhajantri2006camouflage, feng2015camouflage,tankus2001convexity, xue2016camouflage,li2017foreground,pike2018quantifying}. However, these methods usually fell into the trap of camouflage, as the low-level features are often disrupted in camouflage to deceive the perceivers. 
Therefore, recent research usually resorts to the huge capacity of deep network to recognize the 
more complex properties of camouflage.
Among those, Le \etal \cite{le2019anabranch} introduced the joint image classification and camouflaged object segmentation framework.
Yan \etal \cite{yan2020mirrornet} presented an adversarial segmentation stream using a flipped image as input to enhance the discriminative ability of the main segmentation stream for camouflaged object detection. Fan \etal \cite{fan2020camouflaged} proposed SINet to gradually locate and search
for the camouflaged object.
All of the above methods try to mimic the perception and cognition of observers 
performing on camouflaged objects. However, they ignored an important attribute: the time that observers 
spend on searching for the camouflaged object varies in a wide range and heavily depends on the effectiveness of camouflage \cite{troscianko2009camouflage}. Therefore, they fail to consider that the features employed to detect the objects are also different when they have different camouflage degrees, 
which is a useful indicator in camouflage research \cite{background_matching}. To reveal the degree of camouflage, and discover the regions that make camouflaged objects detectable, we introduce the first camouflaged object ranking method and camouflaged object discriminative region localization solution to effectively analyse the attribute of camouflage.

\noindent\textbf{Ranking based dense prediction models:} For some attributes, \eg saliency, it's natural to have ranking in the annotation for better understanding of the task.
Islam \etal \cite{amirul2018revisiting} argued that saliency is a relative concept when multiple observers are queried. Toward this, they collected a saliency ranking dataset based on the PASCAL-S dataset \cite{pascal-s} with 850 images labeled by 12 observers. Based on this dataset, they designed
an encoder-decoder model to predict saliency masks of different levels to achieve the final ranking prediction. Following their idea, Yildirim \etal \cite{yildirim2020evaluating} evaluated saliency ranking based on
the assumption that objects in natural images are perceived to have varying levels of importance.
Siris \etal \cite{siris2020inferring} defined ranking by inferring the order of attention shift when people view
an image. Their dataset is based on the fixation data provided by SALICON \cite{jiang2015salicon}.
As far as we know, there exist no camouflaged object ranking models. Similar to saliency, camouflaged object have levels, and the camouflaged objects of higher level background matching or disruptive coloration may hide better in the environment, indicating a higher level of camouflage. Based on this, our ranking based solution leads to better understanding about evolution of animals. Different from saliency ranking, which
is relative within a single image,
we define camouflage ranking as relative and progressive across the entire dataset, which is generated based on the median fixation time of multiple observers. 



\noindent\textbf{Discriminative region localization technique:} The discriminative regions \cite{zhou2016learning} are those
leading to accurate classification, \eg, the head of the animals
and \etc.
Zhou \etal \cite{zhou2016learning}  introduced the class activation map (CAM) to estimate the discriminative region of each class,
which is the basis of many weakly supervised methods~\cite{ahn2018learning,wei2018revisiting,huang2018weakly,wei2017object,lee2019ficklenet,souly2017semi,wang2017learning}. Selvaraju \etal \cite{selvaraju2017grad} extended CAMs by utilizing the gradient of the class score w.r.t. activation of the last convolutional layer of CNN to investigate the importance of each neuron.
Chattopadhay \etal \cite{chattopadhay2018grad} used a linear combination of positive gradients w.r.t. activation maps of the last convolutional layer to capture the importance of each class activation map for the final classification. Zhang \etal \cite{zhang2018adversarial} erased the high activation area iteratively to force a CNN to learn all relevant features and therefore expanded the discriminative region.  
Similar to the existing discriminative region localization techniques,
we introduce the first camouflaged object
discriminative region localization method
to reveal the most salient region of the camouflaged objects.


\begin{figure*}[!htp]
   \begin{center}
   \begin{tabular}{c@{ }}
   {\includegraphics[width=0.85\linewidth]{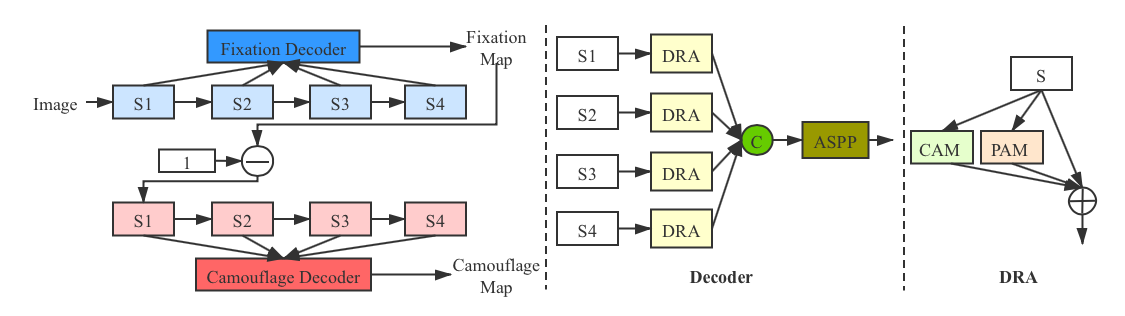}}\\
   \end{tabular}
   \end{center}
  \vspace{-5mm}
   \caption{Overview of the joint fixation and segmentation prediction network. The first part indicates the pipeline that the Fixation Decoder and Camouflage Decoder generates the corresponding maps. The second part is the structrue of the decoders, where \enquote{ASPP} is the denseaspp module \cite{denseaspp}. The third part is the structure of the dual residual attention module \enquote{DRA} in decoder, in which
   \enquote{CAM} and \enquote{PAM} are channel attention module and position attention module from \cite{fu2019dual}.}
  \vspace{-3mm}
\label{fig:joint_cod_fixation}
\end{figure*}

\section{Our Method}
As there exists no localization or ranking based camouflage dataset, we will first discuss our new dataset, and then present our model.


\subsection{The new dataset}
\noindent\textbf{Dataset collection:}
To achieve camouflaged object localization and ranking, we first relabel some images from existing camouflaged object detection datasets CAMO \cite{le2019anabranch} and COD10K \cite{fan2020camouflaged} to have both localization (fixation) annotation and ranking annotation. We denote the reprocessed dataset as CAM-FR.
The basic assumption
is that the longer it takes for the viewer to find the camouflaged object, the higher level of the camouflaged object \cite{troscianko2009camouflage}. Based on this, we record the detection delay for each camouflaged object, and use it as the indicator for camouflage ranking.

To do so, we use an eye tracker (SMI RED250) and record the time for each camouflaged object to be noticed.
SMI RED250 provides three sampling rates, 60Hz, 120Hz and 250Hz, representing the accuracy of the recorded detection delay. We use the 250Hz sampling rate in our experiment. The operating distance is 60-80cm, which is the distance from observers to the camouflaged image. The movement range is 40cm in the horizontal direction and 20cm in the vertical direction, which is the range for the observers to move in order to discover the camouflaged objects.

With the existing camouflaged object detection training datasets, \eg, the COD10K \cite{fan2020camouflaged} and CAMO datasets \cite{le2019anabranch}, we invite six observers to view each image with the task of camouflaged object detection\footnote{We have multiple observers to produce robust level of camouflage}.
We define the median observation time across different observers as the detection delay for each camouflaged instance, with the help of instance-level annotations. 
Specifically, we define the observation time for the $j$-th observer towards the $i$-th instance as:
\begin{equation}
\triangle t_{ij} = \mathrm{median}(\boldsymbol{\delta t_{ij}}), 
\ \boldsymbol{\delta t_{ij}}=\{t_{ij}^{k}-t_{j}^0\}_{k=1}^K
\end{equation}
$K$ is the number of fixation points on the instance, $t_{j}^0$ is the start time for observer $j$ to watch the image and $t_{ij}^{k}$ is the time of the $k$-th fixation point on the instance $i$ with observer $j$. To avoid the influence of extreme high or low fixation time, we use the median instead of the mean value:
\begin{equation}
\mathrm{median}(\boldsymbol{x})=\left\{\begin{array}{cc}x_{(n+1) / 2}, & \mathrm{n} / 2 \neq 0 \\ \frac{x_{\lfloor{n / 2\rfloor}}+x_{\lfloor{n / 2\rfloor}+1}}{2}, & \mathrm{n} / 2=0\end{array}\right.
\end{equation}
in which $\boldsymbol{x}=\{x_l\}_{l=1}^n$ is a set indexed in ascending order.  
Considering different perception ability of observers,
we define the final detection delay for instance $i$ as the median across the six observers: $\triangle t_i = \mathrm{median}_j(\triangle t_{ij})$, which is then used to get our ranking dataset.

There exist two different cases that may result into no fixation points in the camouflaged instance region.
The first is caused by
a mechanical error of the eye tracker or incorrect operation by observers. The second is caused by the higher level of camouflage, which makes it difficult to detect the camouflaged object.
We set a threshold to distinguish these two situations. If more than half of the observers ignore the instance, we consider it as a hard sample and the search time is set to 1 (after normalization). Otherwise, values of the corresponding observers are deleted and the median is computed from the remaining detection decays. 

\noindent\textbf{Dataset information:} Our dataset CAM-FR contains 2,000 images for training and 280 images for testing. The training set includes
1,711 images from the COD10K-CAM training set \cite{fan2020camouflaged} and 289 images are from the CAMO training set \cite{le2019anabranch}. Then, we relabel
238 images from the COD10K-CAM training set and 42 images from the CAMO training set as the testing set. In CAM-FR, we have different ranks (rank 0 is the background), where rank 1 is the hardest level, rank 2 is median
and rank 3 is the easiest level.

\noindent\textbf{Model design with the new dataset:} Based on our new dataset, we propose to simultaneously localize, segment and rank the camouflaged objects. Given an input image, the first two tasks regress the fixation map and segmentation map respectively, while the third task involves instance segmentation (camouflaged object detection) and classification (camouflaged object ranking). We build the three tasks within one unified framework as shown in Fig.~\ref{fig:network_overview}, where the localization network and segmentation network are integrated in one joint learning framework. The ranking model shares the backbone network with the joint learning framework to produce camouflage ranking.

\subsection{Joint localization and segmentation}
\noindent\textbf{Task analysis:}
We define the \enquote{discriminative region} as a region that makes the camouflaged object apparent. Compared with other regions of the camouflaged object, the discriminative region should have a higher contrast with it's surroundings than the other regions of the camouflaged object.
Based on this observation, we design a reverse attention module
based joint camouflaged object discriminative region localization and segmentation network in Fig.~\ref{fig:joint_cod_fixation}, which can simultaneously regress the discriminative regions that make the camouflaged objects obvious and segment the camouflaged objects.

\noindent\textbf{Network design:}
We built our joint learning framework with ResNet50 \cite{he2016deep} as backbone shown in Fig.~\ref{fig:joint_cod_fixation}. Given an input image $I$, we feed it to the backbone to obtain feature representation $s_1,s_2,s_3,s_4$, representing feature maps from different stages of the backbone network. Similar to existing ResNet50 based networks, we define a group of convolutional layers that produce the same spatial size as belonging to
the same stage of the network. Then we design the \enquote{Fixation Decoder} and \enquote{Camouflage Decoder} modules with the same network structure, as \enquote{Decoder} in Fig.~\ref{fig:joint_cod_fixation}, to regress the fixation map and segmentation map respectively. Each $s_i$, $i=1,...,4$ is fed to a convolutional layer of kernel size $3\times3$ to achieve the new feature map $\{s_i'\}_{i=1}^4$ of channel dimension $C=32$ respectively. Then, we propose the dual residual attention model as \enquote{DRA} in Fig.~\ref{fig:joint_cod_fixation}  by modifying the dual attention module \cite{fu2019dual}, to obtain a discriminative feature representation
with a position attention module (PAM) and channel attention module (CAM). The \enquote{ASPP} in the decoder is the denseaspp module in \cite{denseaspp} to achieve a multi-scale receptive field.

With the proposed \enquote{Fixation Decoder} module, we obtain our discriminative region, which will be compared with the provided ground truth fixation map to produce our loss function for the fixation branch. Then, based on our observation that the fixated region usually has higher saliency than the other parts of the object, we introduce a reverse attention based framework to jointly learn the discriminative region and regress the whole camouflaged object.
Specifically, given the discriminative region prediction $F$, we obtain the reverse attention as $1-F$. Then we treat it as the attention 
and multiply it with the backbone feature $s_1$ to generate the reverse attention guided feature $\{s_i^r\}_{i=1}^4$ similar to \cite{cpd_sal}. Then, we have the \enquote{Camouflage Decoder} to generate our saliency prediction $S$ from $\{s_i^r\}_{i=1}^4$.

\noindent\textbf{Objective function:}
We have two loss functions in the joint learning framework: the discriminative region localization loss and the camouflaged object detection loss. For the former, we use the
binary cross-entropy loss $\mathcal{L}_f$. For the latter, we adopt the
pixel position aware loss as in \cite{wei2020f3net} to produce predictions with higher structure accuracy. Then we define our joint learning framework based loss function as:
\begin{equation}
\label{joint_loss}
\mathcal{L}_{fc} = \mathcal{L}_f + \lambda \mathcal{L}_c,
\end{equation}
where $\lambda$ is a weight to measure the importance of each task, and empirically we set $\lambda=1$ in this paper.

\subsection{Inferring the ranks of camouflaged objects}
\noindent\textbf{Instance segmentation based rank model:}
We construct our camouflage ranking model on the basis of Mask R-CNN \cite{he2017mask} to learn the degree of camouflage.
Similar to the goal of Mask R-CNN \cite{he2017mask}, the aim of the camouflage ranking model is jointly segmenting the camouflaged objects and inferring their ranks. Following the standard pipeline of Mask R-CNN, we design a camouflaged object ranking model as shown in Fig.~\ref{fig:network_overview}, with the \enquote{Instance Segmentation} branch supervised by the binary ground truth of each camouflaged instance, and an \enquote{Instance Classification} branch to produce the camouflaged object ranking.


Firstly, we feed the image $I\in \mathbb{R}^{h\times w\times 3}$ into the backbone network (ResNet50 \cite{he2016deep} in particular)
to extract image features.
Then the \enquote{Feature Pyramid Network} (FPN) \cite{fpn} is employed to integrate the feature maps of different levels. The final set of feature maps is denoted as $P=\{P_1, \cdots, P_n\}$, where $n$ is the number of layers. Then the \enquote{Region Proposal Network} (RPN) \cite{faster_rcnn} is adopted, which takes the feature of the whole image as input, and detects the regions that are likely to contain the camouflaged instances, \ie the regions of interest (ROIs). Two branches are included in RPN: 1)
a classification branch, which determines whether the candidate bounding box contains the camouflaged object; and 2) a regression branch, which regresses the coordinates of the ground truth camouflaged object bounding box. 

With features produced by FPN, the ROIAlign module \cite{he2017mask} is used to extract feature maps of the ROIs. Then, we
predict the rank and regress the location of the camouflaged object, respectively.
Finally, features of the detected camouflaged object are
fed into a segmentation branch to output a binary mask for each camouflaged instance.

During training, a multi-task loss with three components is minimized:
\begin{equation}
\label{mask_rcnn_loss}
    \mathcal{L} = \mathcal{L}_{rpn}+\mathcal{L}_{rank}+\mathcal{L}_{mask},
\end{equation}
where $\mathcal{L}_{rpn}$ is to train the RPN, $\mathcal{L}_{rank}$ is the loss for the ranking model, and $\mathcal{L}_{mask}$ is only defined on the region where the prediction of rank is not 0 (background) and allows the network to segment instances of each rank. Both $\mathcal{L}_{rpn}$ and $\mathcal{L}_{rank}$ consist of classification loss and regression loss. For RPN, it aims to check the existence of the camouflaged instance in the proposal and regress its location. For the rank model, it infers the rank of camouflage and regresses object location.

\begin{figure}[!htp]
   \begin{center}
   \begin{tabular}{c@{ }}
   {\includegraphics[width=0.70\linewidth]{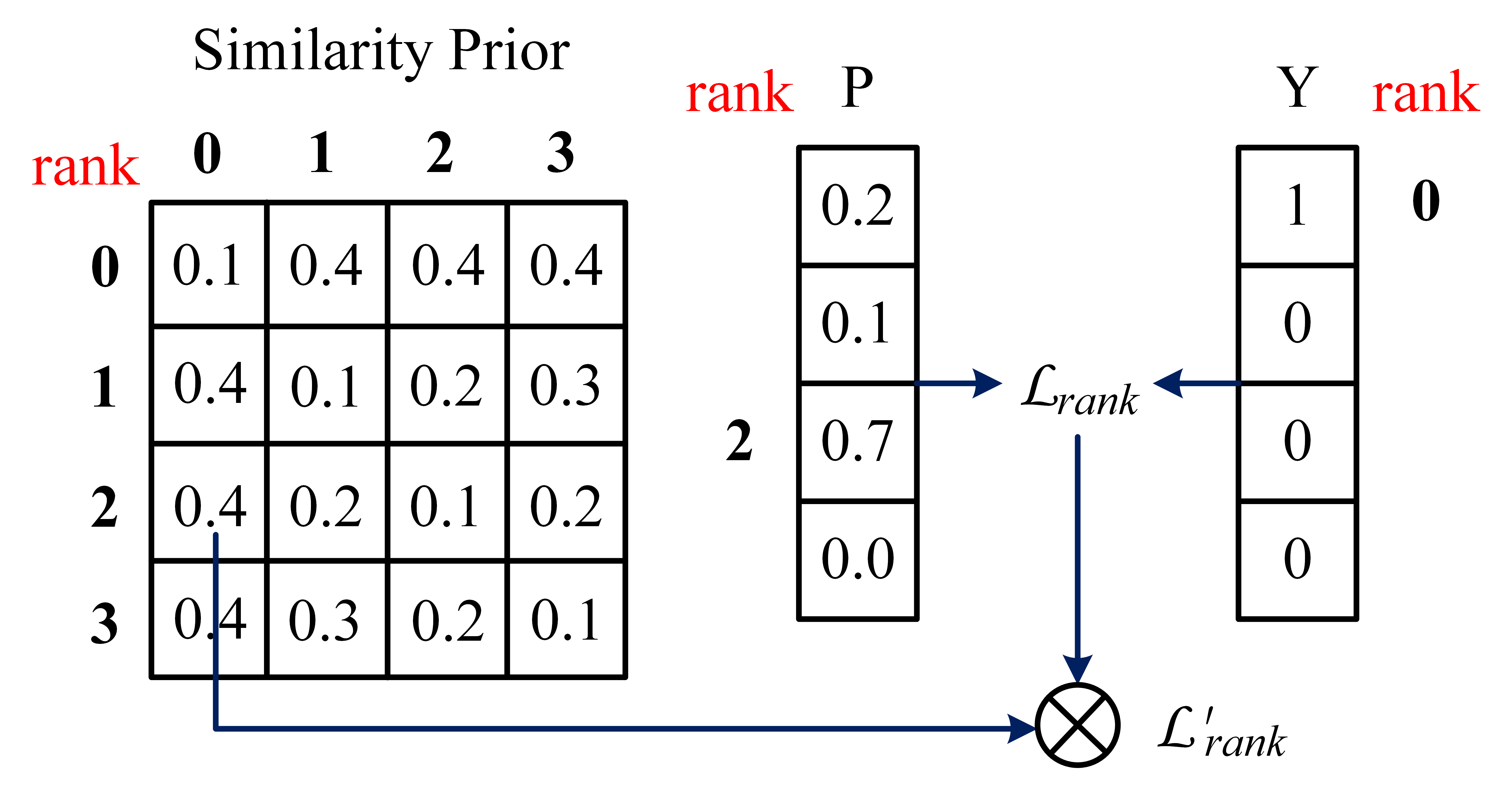}}\\
   \end{tabular}
   \end{center}
  \vspace{-7mm}
   \caption{Label similarity as a prior to consider the rank label dependency of our ranking dataset. $\mathrm{P}$ and $\mathrm{Y}$ denote the prediction and the one-hot ground truth, respectively.
   }
  \vspace{-3mm}
\label{fig:similarity_prior}
\end{figure}

\noindent\textbf{Label similarity as prior:}
Directly inferring ranks of camouflage with Mask-RCNN may produce unsatisfactory results due to the independence of labels in the instance segmentation dataset.
However, in our ranking scenario, the ranks are progressive, \eg camouflaged object of rank 3 (the easiest level) is easier to notice than rank 2 (the median).
Moreover, the instance of rank 1 should be penalized more if it's misclassified as rank 3 instead of rank 2. Towards this, we intend to employ such a constraint on $\mathcal{L}_{rank}$ in Eq.~\ref{mask_rcnn_loss}. Specifically, we define a camouflaged instance similarity prior $S_p$, which is a $4\times4$ matrix as shown in Fig.~\ref{fig:similarity_prior}, with 
each $S_p(m,n)$ representing the penalty for
predicting rank $n$ as rank $m$. Given the prediction of the instance classification network in Fig.~\ref{fig:network_overview}, and the ground truth instance rank, we first compute the original rank loss $\mathcal{L}_{rank}$ (before we compute the mean of $\mathcal{L}_{rank}$). Then, we weight it with the specific similarity prior $S_p(m,n)$. As is illustrated in Fig.~\ref{fig:similarity_prior}, the predicted rank is 2, and the ground truth rank is 0, then we get penalty $S_p(2,0)=0.4$, and multiply it with the original rank loss $\mathcal{L}_{rank}$ to obtain the weighted loss $\mathcal{L}'_{rank}$. Although we pay more attention to misclassified samples, a weight should be assigned to the loss of correct samples, making them produce more confident scores.

\begin{table*}[t!]
  \centering
  \scriptsize
  \renewcommand{\arraystretch}{1.0}
  \renewcommand{\tabcolsep}{1.0mm}
  \caption{Performance of baseline models trained with our CAM-FR dataset on benchmark testing sets.}
  \begin{tabular}{l|cccc|cccc|cccc|cccc}
  \hline
  &\multicolumn{4}{c|}{CAMO}&\multicolumn{4}{c|}{CHAMELEON}&\multicolumn{4}{c|}{COD10K}&\multicolumn{4}{c}{NC4K} \\
    Method & $S_{\alpha}\uparrow$&$F^{\mathrm{mean}}_{\beta}\uparrow$&$E^{\mathrm{mean}}_{\xi}\uparrow$&$\mathcal{M}\downarrow$& $S_{\alpha}\uparrow$&$F^{\mathrm{mean}}_{\beta}\uparrow$&$E^{\mathrm{mean}}_{\xi}\uparrow$&$\mathcal{M}\downarrow$ &  $S_{\alpha}\uparrow$ & $F^{\mathrm{mean}}_{\beta}\uparrow$ & $E^{\mathrm{mean}}_{\xi}\uparrow$ & $\mathcal{M}\downarrow$ & $S_{\alpha}\uparrow$
    & $F^{\mathrm{mean}}_{\beta}\uparrow$ & $E^{\mathrm{mean}}_{\xi}\uparrow$ & $\mathcal{M}\downarrow$  \\
  \hline
  SCRN \cite{scrn_sal}& 0.702 & 0.632 & 0.731 & 0.106 & 0.822 & 0.726 & 0.833 & 0.060 & 0.756 & 0.623 & 0.793 & 0.052  & 0.793 & 0.729 & 0.823 & 0.068  \\
  CSNet\cite{csnet_eccv} & 0.704 & 0.633 & 0.753 & 0.106 & 0.819 & 0.759 & 0.859 & 0.051 & 0.745 & 0.615 & 0.808 & 0.048  & 0.785 & 0.729 & 0.834 & 0.065  \\ 
  UCNet \cite{ucnet_sal} & 0.703 & 0.640 & 0.740 & 0.107 & 0.833 & 0.781 & 0.890 & 0.049 & 0.756 & 0.650 & 0.823 & 0.047  & 0.792 & 0.751 & 0.854 & 0.065  \\ 
  BASNet \cite{basnet_sal} & 0.644 & 0.578 & 0.588 & 0.143 & 0.761 & 0.657 & 0.797& 0.080 & 0.640 & 0.579 & 0.713 & 0.072  & 0.724 & 0.648 & 0.780 & 0.089  \\
  SINet \cite{fan2020camouflaged} & 0.697 & 0.579 & 0.693 & 0.130 & 0.820 & 0.731 & 0.835 & 0.069 & 0.733 & 0.588 & 0.768 & 0.069  & 0.779 & 0.696 & 0.800 & 0.086  \\ \hline
  Ours\_cod\_new  & \textbf{0.708} & \textbf{0.645} & \textbf{0.755} & \textbf{0.105} & \textbf{0.842} & \textbf{0.794} & \textbf{0.896} & \textbf{0.046} & \textbf{0.760} & \textbf{0.658} & \textbf{0.831} & \textbf{0.045}  & \textbf{0.797} & \textbf{0.758} & \textbf{0.854} & \textbf{0.061}  \\ 
   \hline
  \end{tabular}
  \label{tab:benchmark_model_comparison_rankingg_dataset}
\end{table*}

\section{Experimental Results}
\subsection{Setup}
\noindent\textbf{Dataset:}
We train our framework with CAM-FR training set
to provide a baseline to simultaneously achieve camouflaged object detection (Ours\_cod\_new), discriminative region localization (Ours\_fix\_new) and camouflage ranking (Ours\_rank\_new) and test on the testing set of CAM-FR.
Further, to compare our performance with benchmark
models, we further train a single camouflaged object detection model (Ours\_cod\_full) with the conventional training dataset which contains 3,040 images from COD10K and 1,000 images from CAMO, and test on the existing testing datasets, including
CAMO \cite{le2019anabranch}, COD10K \cite{fan2020camouflaged}, CHAMELEMON \cite{Chameleon2018} and our NC4K testing dataset.

\noindent\textbf{Training details:}
A pretrained ResNet50 \cite{he2016deep} is employed as our backbone network.
During training, the input image is resized to $352\times 352$. Candidate bounding boxes spanning three scales (4, 8, 16) and three aspect ratios (0.5, 1.0, 2.0) are selected from each pixel. 
In the RPN module of the ranking model, the IoU threshold with the ground truth is set to 0.7, which is used to determine whether the candidate bounding box is positive (IoU$>$0.7) or negative (IoU$<$0.7) in the next detection phase. The IoU threshold is set to 0.5 to determine whether the camouflaged instances are detected and only positive ones are sent into the segmentation branch. Our model in Fig.~\ref{fig:network_overview} is trained on one GPU (Nvidia RTX 1080 Ti) for 10k iterations (14 hours) with a mini-batch of 10 images, using the Adam optimizer with a learning rate of 5e-5. 

\begin{figure*}[!htp]
   \begin{center}
   \begin{tabular}{c@{ } c@{ } c@{ } c@{ } c@{ } c@{ }}
      {\includegraphics[width=0.15\linewidth, height=0.125\linewidth]{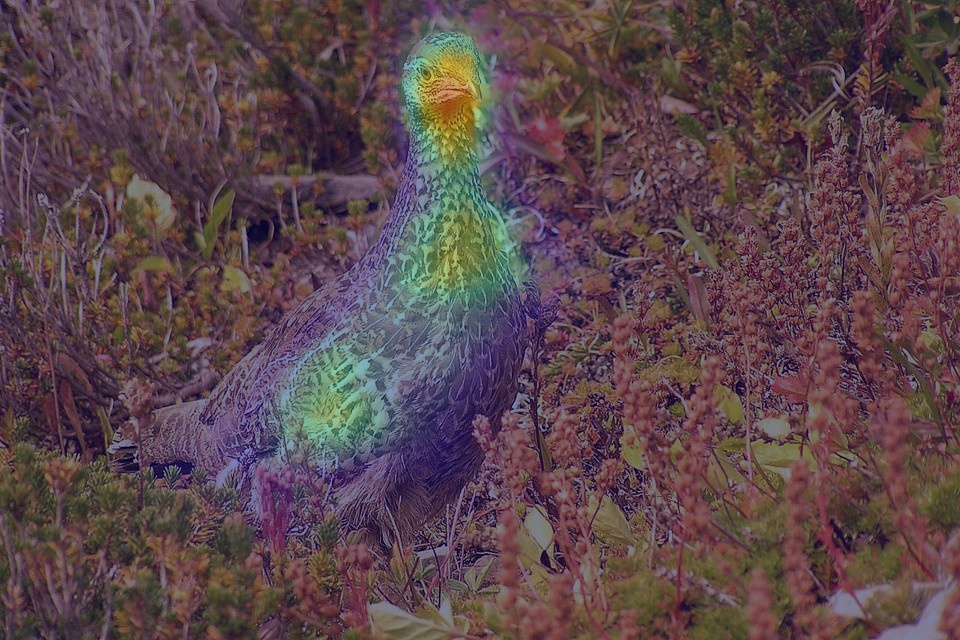}}&
    {\includegraphics[width=0.15\linewidth,height=0.125\linewidth]{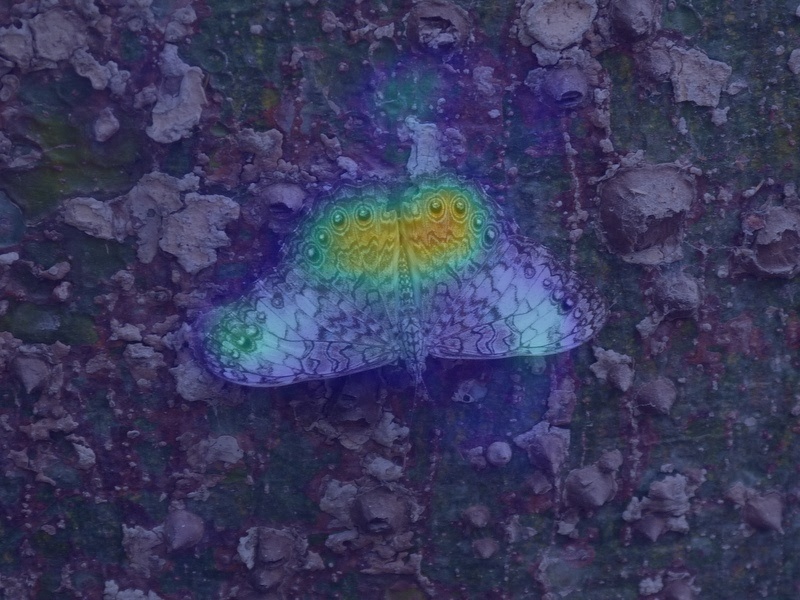}}&
    {\includegraphics[width=0.15\linewidth,height=0.125\linewidth]{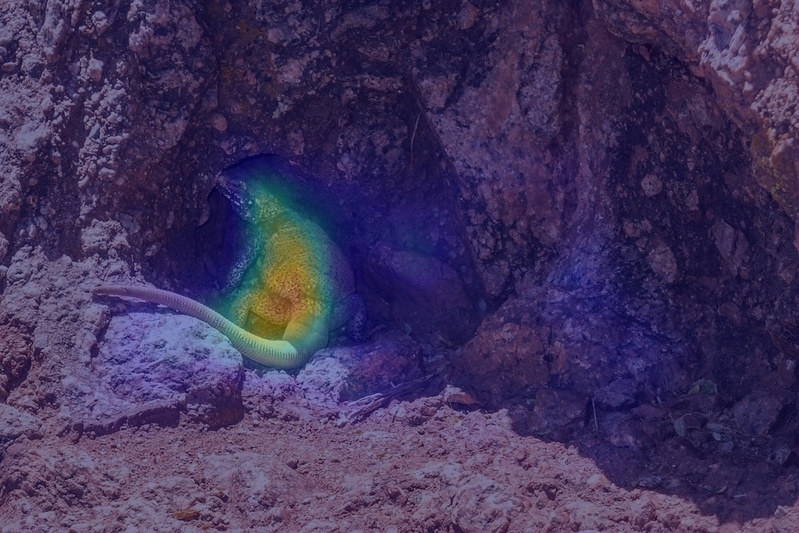}}&
    {\includegraphics[width=0.15\linewidth,height=0.125\linewidth]{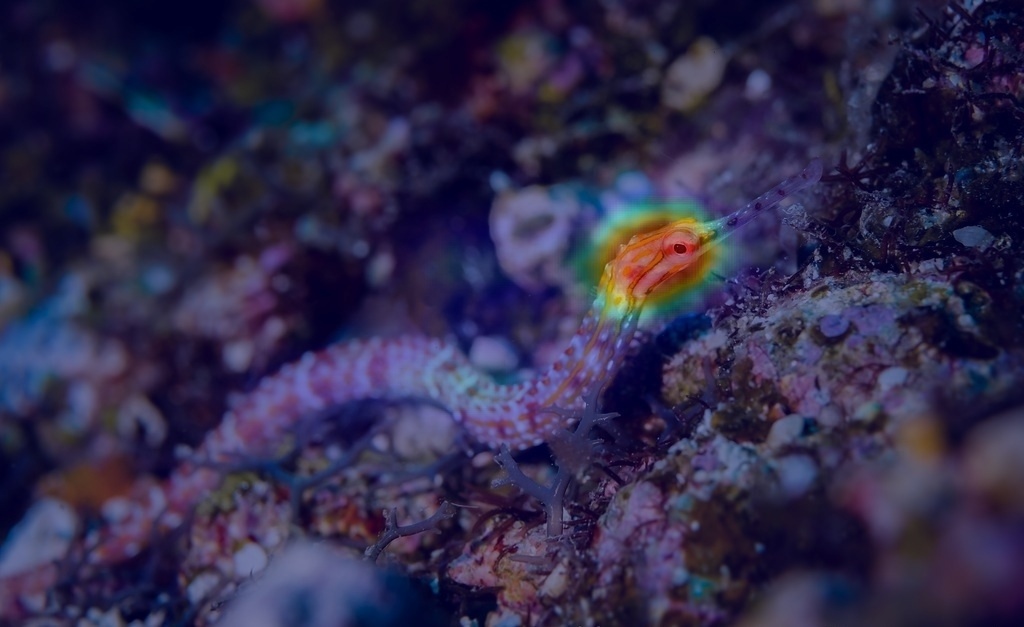}}&
    {\includegraphics[width=0.15\linewidth,height=0.125\linewidth]{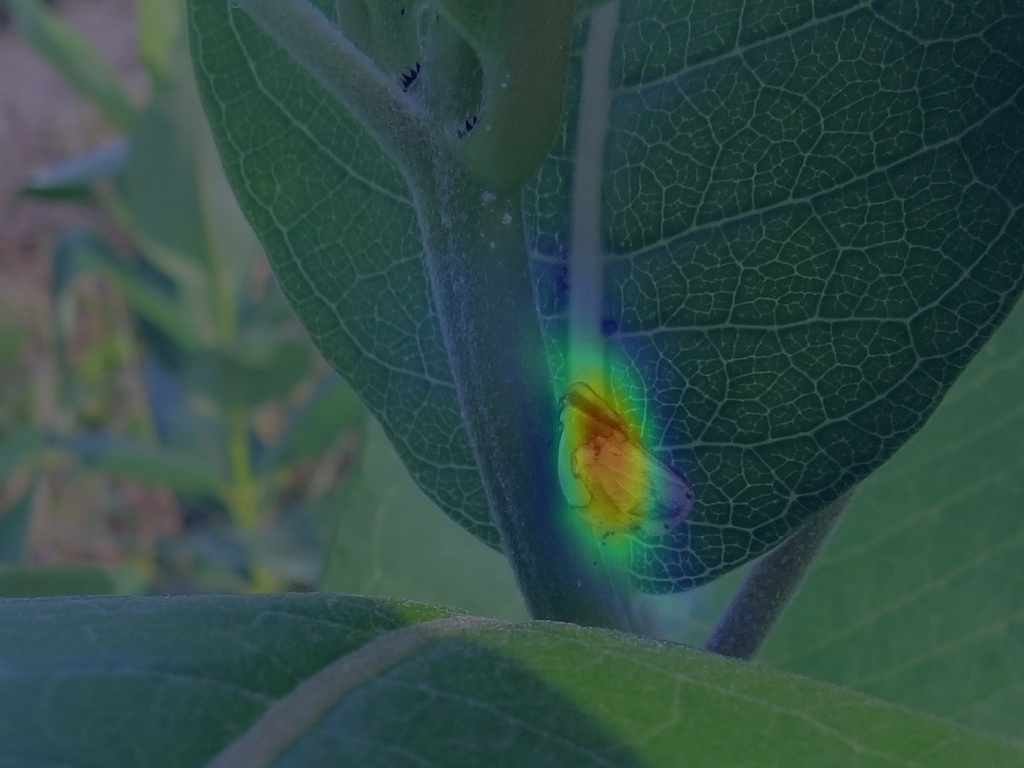}}&
    {\includegraphics[width=0.15\linewidth,height=0.125\linewidth]{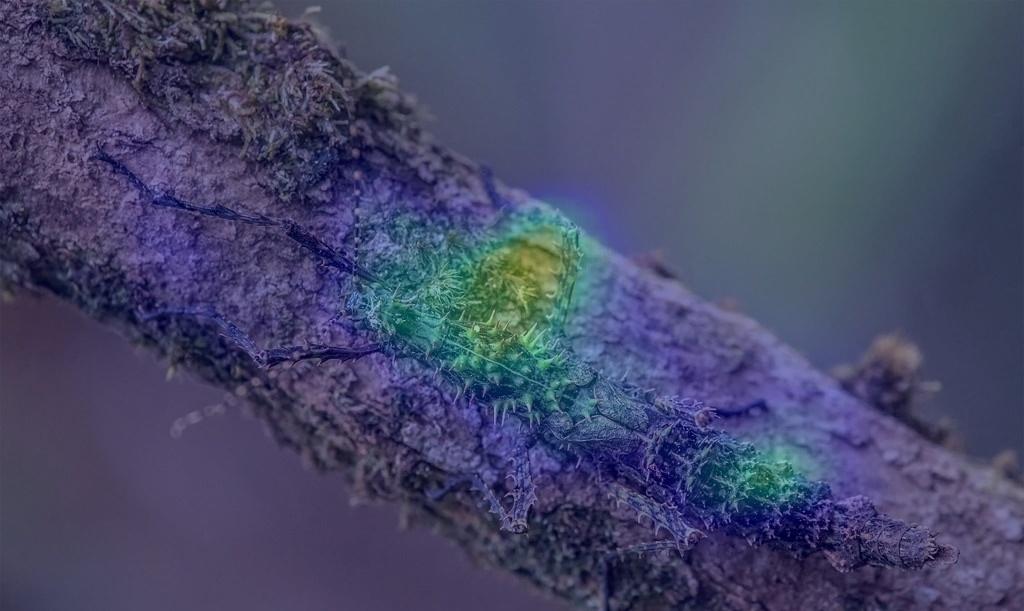}}\\
   {\includegraphics[width=0.15\linewidth, height=0.125\linewidth]{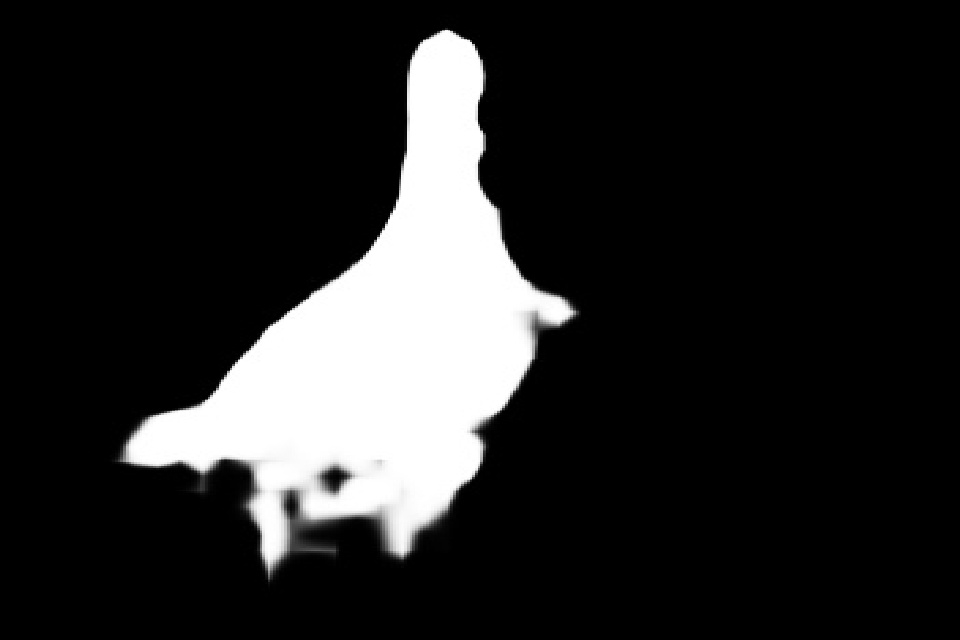}}&
    {\includegraphics[width=0.15\linewidth,height=0.125\linewidth]{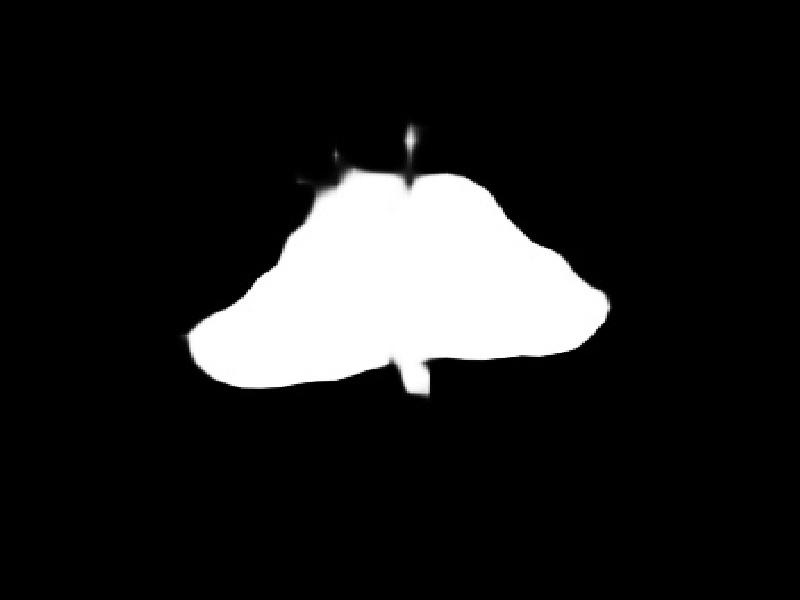}}&
    {\includegraphics[width=0.15\linewidth,height=0.125\linewidth]{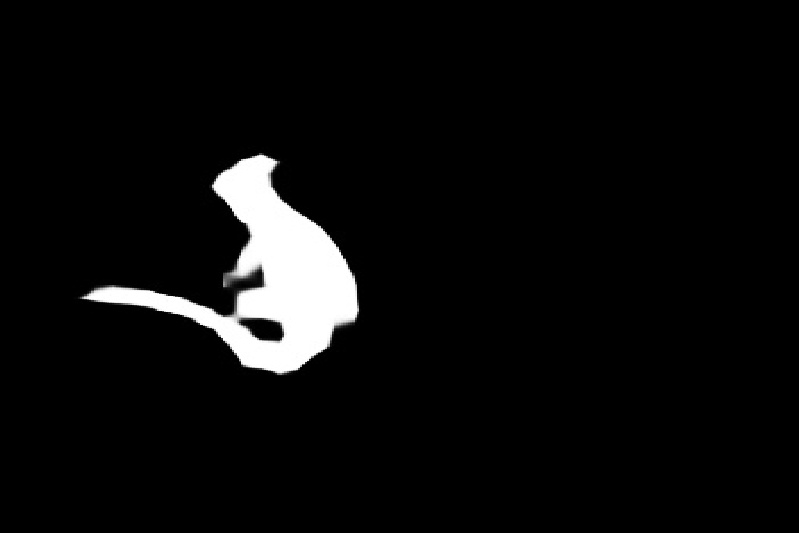}}&
    {\includegraphics[width=0.15\linewidth,height=0.125\linewidth]{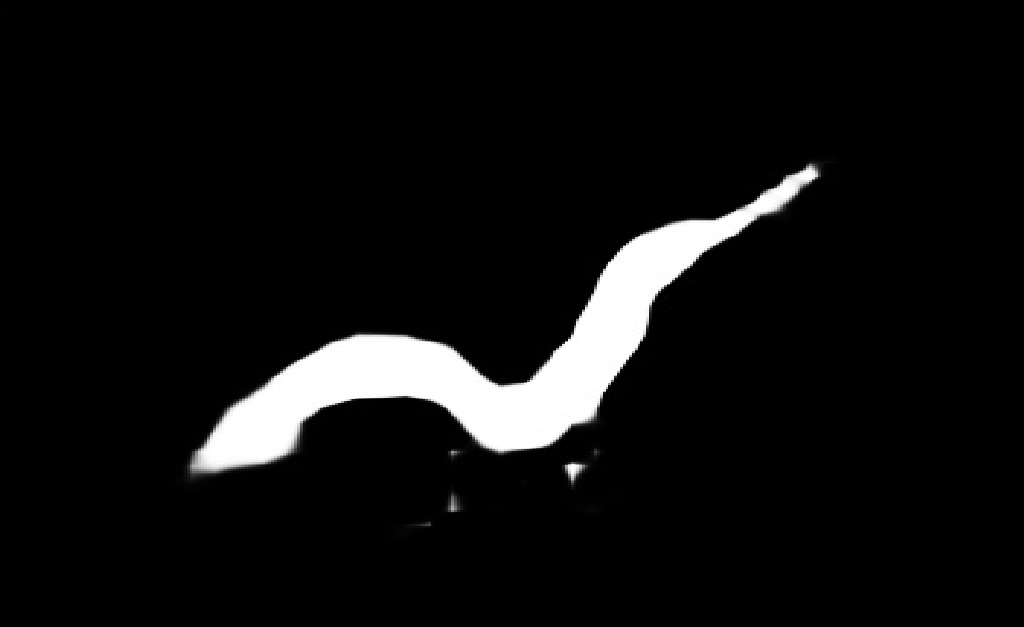}}&
    {\includegraphics[width=0.15\linewidth,height=0.125\linewidth]{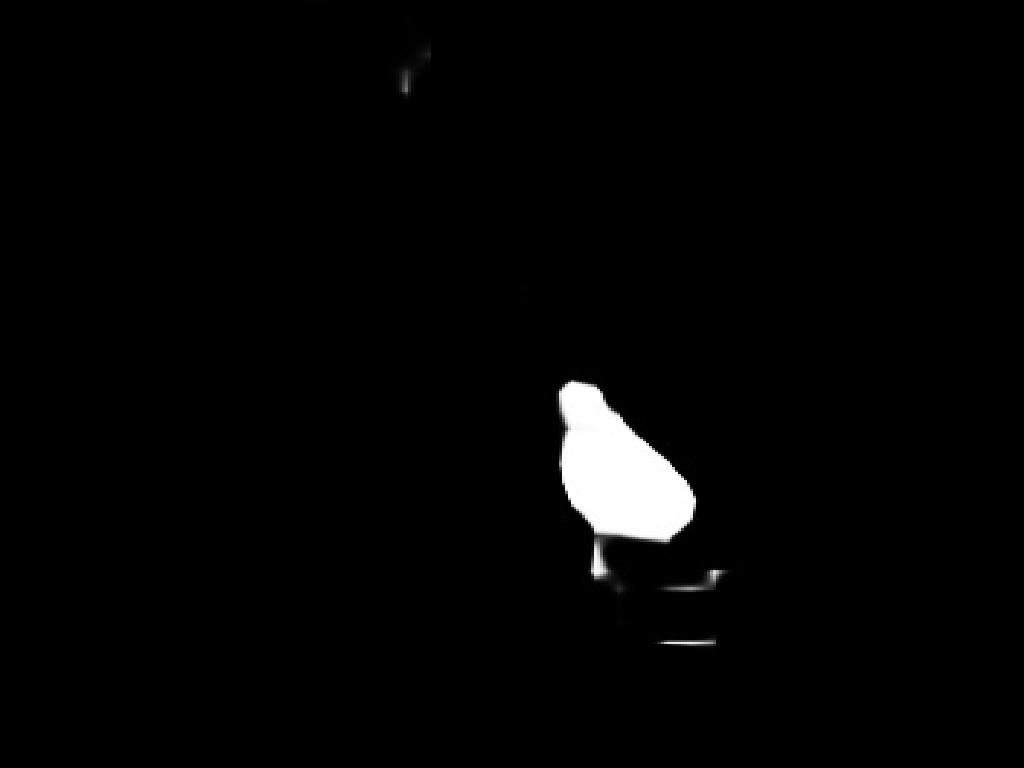}}&
    {\includegraphics[width=0.15\linewidth,height=0.125\linewidth]{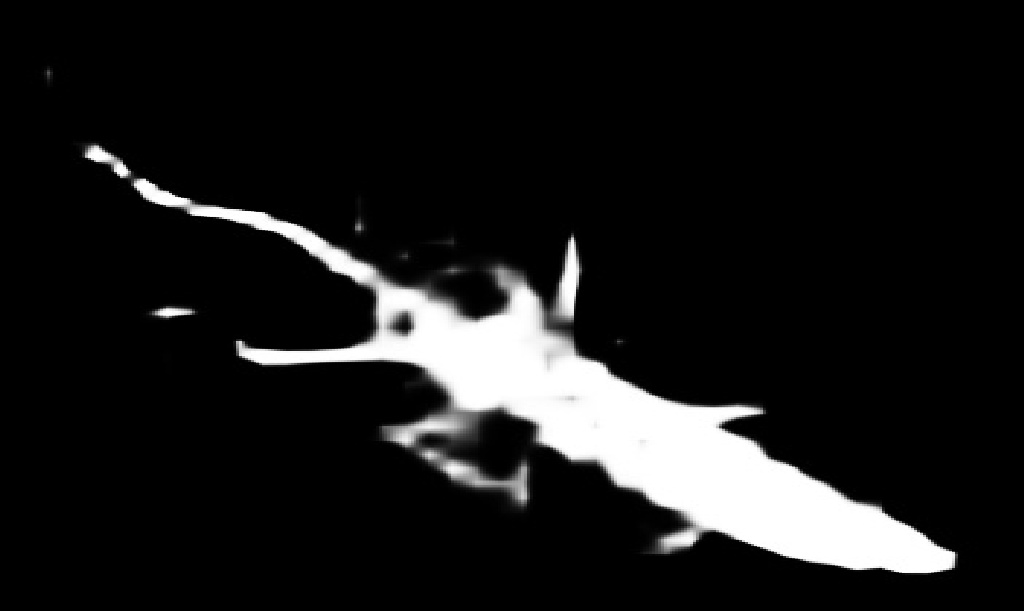}}\\
    {\includegraphics[width=0.15\linewidth, height=0.125\linewidth]{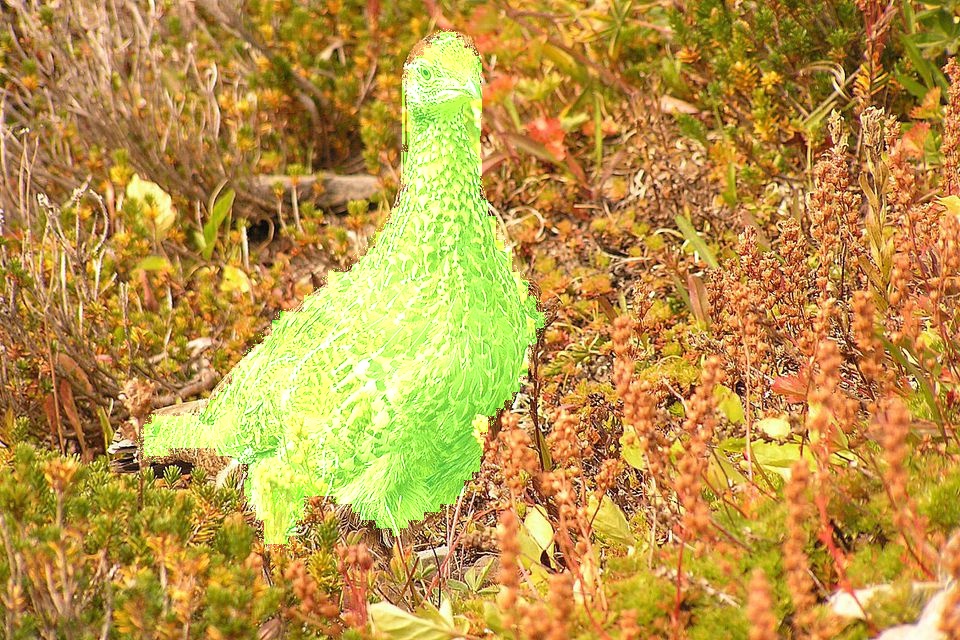}}&
   {\includegraphics[width=0.15\linewidth, height=0.125\linewidth]{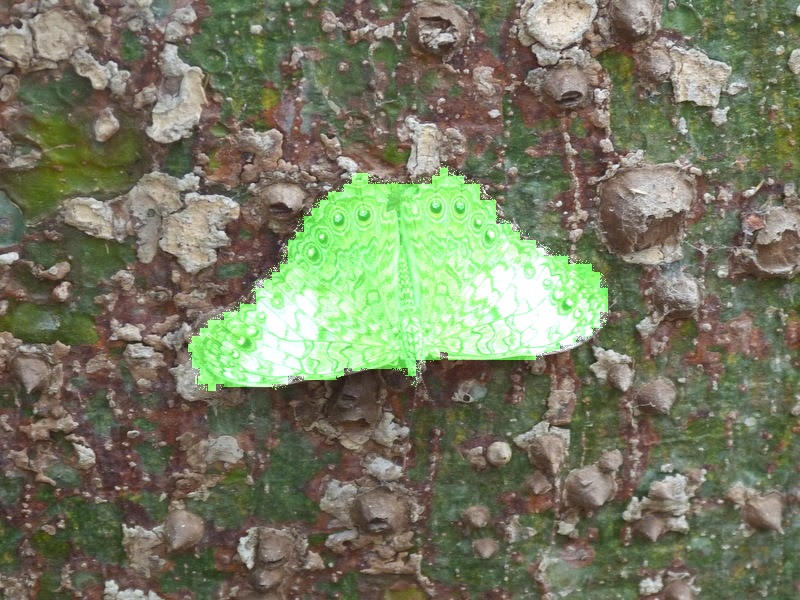}}&
    {\includegraphics[width=0.15\linewidth,height=0.125\linewidth]{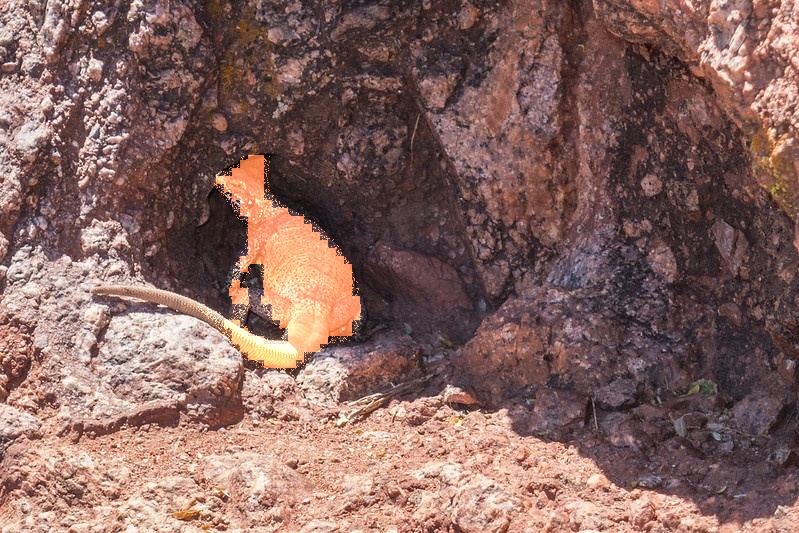}}&
    {\includegraphics[width=0.15\linewidth,height=0.125\linewidth]{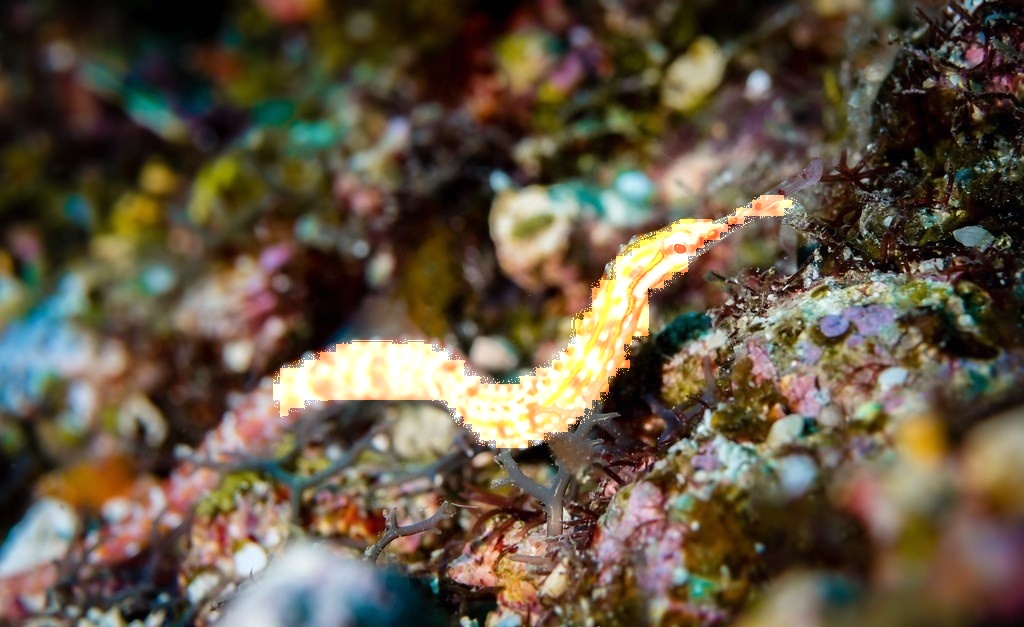}}&
    {\includegraphics[width=0.15\linewidth,height=0.125\linewidth]{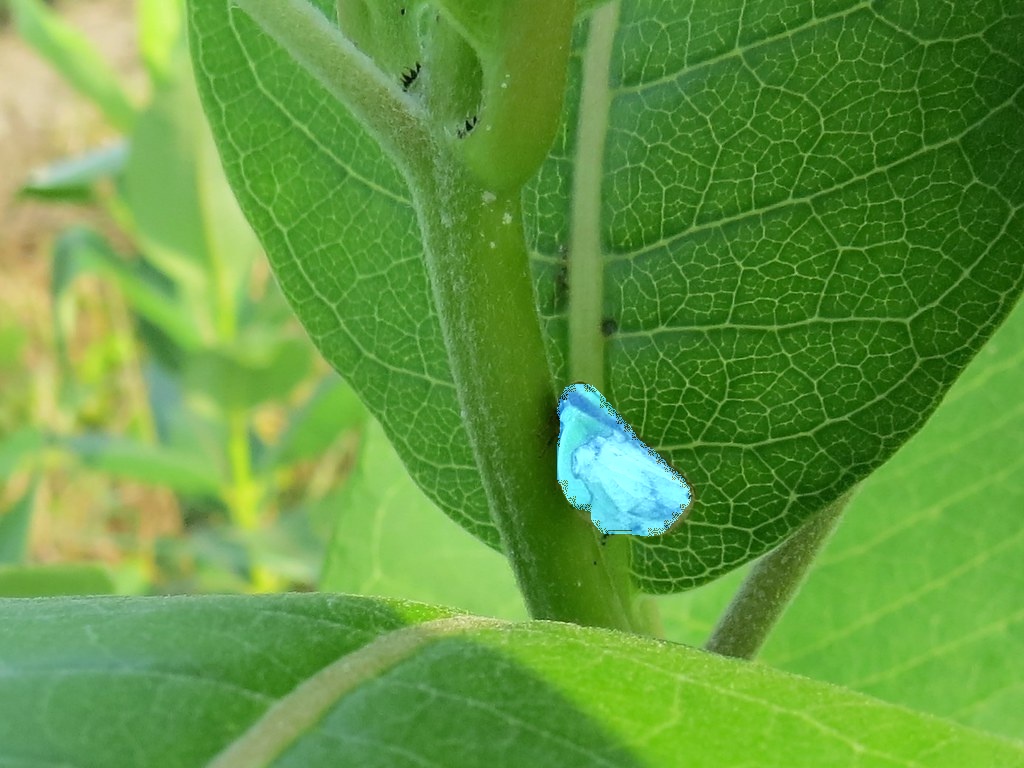}}&
    {\includegraphics[width=0.15\linewidth,height=0.125\linewidth]{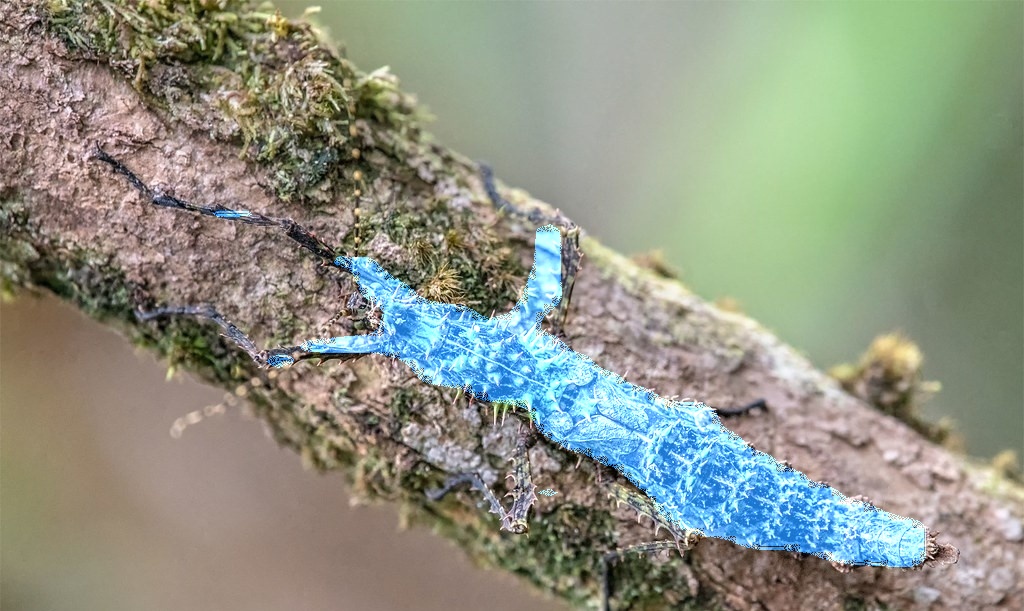}}\\
   \end{tabular}
   \end{center}
   \vspace{-20pt}
   \caption{Visual performance of the proposed algorithm. From top to bottom: fixation, camouflaged object detection and ranking predictions. The green, orange and blue annotations in the third row indicate camouflage rank 3 (easiest), 2 (median) and 1 (hardest), respectively. }
  \vspace{-3mm}
\label{fig:predictions_visualization}
\end{figure*}

\noindent\textbf{Evaluation metrics:}
Conventionally, camouflaged object detection is defined as a binary segmentation task, and the widely used evaluation metrics include Mean Absolute Error, Mean F-measure, Mean E-measure \cite{fan2018enhanced} and S-measure \cite{fan2017structure} denoted as $\mathcal{M}$, $F_\beta^{\mathrm{mean}}$, $E_\xi^{\mathrm{mean}}$, $S_{\alpha}$, respectively.


We find that the above four evaluation metrics cannot evaluate the performance of ranking based prediction.
For the ranking task, \cite{amirul2018revisiting} introduced the Salient Object Ranking (SOR) metric to measure ranking performance, which is defined as the Spearman’s Rank-Order Correlation between the ground truth rank order and the predicted rank order of salient objects. However, it cannot be used in our scenario, as
Spearman’s Rank-Order Correlation is based on at least two different ranking levels. However, in our ranking based dataset, most of the images have only one camouflaged object. To deal with this, we introduce $r_{MAE}$:
\begin{equation}
    r_{MAE}=\frac{\sum_{i=1}^w\sum_{j=1}^h|r_{ij}-\hat{r}_{ij}|}{N},
\end{equation}
where $N$ is the number of pixels, $w$ and $h$ are the width and height of the image. $\hat{r}_{ij}$ and $r_{ij}$ are the predicted and ground truth ranks respectively with values $0,1,2,3$ corresponding to \enquote{background}, \enquote{hardest}, \enquote{median} and \enquote{easiest}, respectively. If the prediction is consistent with the ground truth, their difference is supposed to be 0. In $r_{MAE}$, an \enquote{easiest} sample is
punished less when it is predicted as a \enquote{median} sample than as a \enquote{hardest} sample. Accordingly, it is a convincing metric to evaluate the performance of ranking.
For the discriminative region localization, we adopt the widely used fixation prediction evaluation metrics including Similarity ($SIM$) \cite{judd2012benchmark}, Linear Correlation Coefficient ($CC$) \cite{le2007predicting}, Earth Mover’s Distance ($EMD$)  \cite{rubner2000earth}, Kullback–Leibler Divergence ($KLD$) \cite{kullback1951information}, Normalized Scanpath Saliency ($NSS$) \cite{peters2005components}, AUC\_Judd ($AUC\_J$) \cite{judd2009learning}, AUC\_Borij ($AUC\_B$)\cite{borji2013analysis}, shuffled AUC ($sAUC$) \cite{borji2012quantitative} as shown in Table~\ref{tab:fixation_baseline}.

\begin{table}[t!]
  \centering
  \scriptsize
  \renewcommand{\arraystretch}{1.05}
  \renewcommand{\tabcolsep}{0.45mm}
  \caption{Performance of the discriminative region localization.}
  \begin{tabular}{cccccccc}
  \hline
  $SIM\uparrow$ & $CC\uparrow$ & $EMD\downarrow$ & $KLD\downarrow$ & $NSS\uparrow$ & $AUC\_J\uparrow$ & $AUC\_B\uparrow$ & $sAUC\uparrow$  \\ \hline
   0.622 & 0.776 & 3.361 & 0.995 & 2.608 & 0.901 & 0.844 & 0.658  \\
   \hline
  \end{tabular}
  \label{tab:fixation_baseline}
\end{table}

\begin{table}[t!]
  \centering
  \scriptsize
  \renewcommand{\arraystretch}{1.0}
  \renewcommand{\tabcolsep}{8.5mm}
  \caption{Comparison of camouflage ranking methods.}
  \begin{tabular}{r|cccc}
  \hline
  Method & $MAE$ & $r_{MAE}$ \\\hline
  Ours\_rank\_new & \textbf{0.049} & \textbf{0.139} \\
  SOLOv2\cite{wang2020SOLOv2} & \textbf{0.049} & 0.210 \\
  MS-RCNN\cite{liu2018path} & 0.053 & 0.142 \\
  RSDNet\cite{amirul2018revisiting} & 0.074 & 0.293 \\
   \hline
  \end{tabular}
  \label{tab:ranking_comparison}
  \vspace{-5mm}
\end{table}

\begin{table*}[t!]
  \centering
  \scriptsize
  \renewcommand{\arraystretch}{1.0}
  \renewcommand{\tabcolsep}{1.0mm}
  \caption{Performance comparison with baseline models on benchmark dataset and our NC4K dataset.}
  \begin{tabular}{l|cccc|cccc|cccc|cccc}
  \hline
  &\multicolumn{4}{c|}{CAMO}&\multicolumn{4}{c|}{CHAMELEON}&\multicolumn{4}{c|}{COD10K}&\multicolumn{4}{c}{NC4K} \\
    Method & $S_{\alpha}\uparrow$&$F_{\beta}^{\mathrm{mean}}\uparrow$&$E_{\xi}^{\mathrm{mean}}\uparrow$&$\mathcal{M}\downarrow$& $S_{\alpha}\uparrow$&$F_{\beta}^{\mathrm{mean}}\uparrow$&$E_{\xi}^{\mathrm{mean}}\uparrow$&$\mathcal{M}\downarrow$ &  $S_{\alpha}\uparrow$ & $F_{\beta}^{\mathrm{mean}}\uparrow$ & $E_{\xi}^{\mathrm{mean}}\uparrow$ & $\mathcal{M}\downarrow$ & $S_{\alpha}\uparrow$
    & $F_{\beta}^{\mathbf{\mathrm{mean}}}\uparrow$ & $E_{\xi}^{\mathbf{\mathrm{mean}}}\uparrow$ & $\mathcal{M}\downarrow$  \\
  \hline
  PiCANet\cite{picanet} & 0.701 & 0.573 & 0.716 & 0.125 & 0.765 & 0.618 & 0.779 & 0.085 & 0.696 & 0.489 & 0.712 & 0.081 & 0.758 & 0.639 & 0.773 & 0.088 \\ 
  CPD \cite{cpd_sal} & 0.716 & 0.618 & 0.723 & 0.113 & 0.857 & 0.771 & 0.874 & 0.048 & 0.750 & 0.595 & 0.776 & 0.053 & 0.790 & 0.708 & 0.810 & 0.071  \\
  SCRN \cite{scrn_sal}& 0.779 & 0.705 & 0.796 & 0.090 & 0.876 & 0.787 & 0.889 & 0.042 & 0.789 & 0.651  & 0.817 & 0.047 & 0.832 & 0.759 & 0.855 & 0.059  \\
  CSNet\cite{csnet_eccv} & 0.771 & 0.705 & 0.795 & 0.092 & 0.856 & 0.766 & 0.869 & 0.047 & 0.778 & 0.635 & 0.810 & 0.047  & 0.819 & 0.748 & 0.845 & 0.061  \\ 
  PoolNet \cite{Liu19PoolNet}  & 0.730 & 0.643 & 0.746 & 0.105 & 0.845 & 0.749 & 0.864 & 0.054 & 0.740 & 0.576 & 0.776 & 0.056  & 0.785 & 0.699 & 0.814 & 0.073  \\
  UCNet \cite{ucnet_sal} & 0.739 & 0.700 & 0.787 & 0.094 & 0.880 & 0.836 & 0.930 & 0.036 & 0.776 &0.681 & 0.857 & 0.042  & 0.813 & 0.777 & 0.872 & 0.055  \\ 
  F3Net \cite{wei2020f3net} & 0.711 & 0.616 & 0.741 & 0.109 & 0.848 & 0.770 & 0.894 & 0.047 & 0.739 & 0.593 & 0.795 & 0.051 & 0.782 & 0.706 & 0.825 & 0.069 \\
  ITSD\cite{zhou2020interactive} & 0.750 & 0.663 & 0.779 & 0.102 & 0.814 & 0.705 & 0.844 & 0.057 & 0.767 & 0.615 & 0.808 & 0.051  & 0.811 & 0.729 & 0.845 & 0.064  \\ 
  BASNet \cite{basnet_sal} & 0.615 & 0.503 & 0.671 & 0.124 & 0.847 & 0.795 & 0.883 & 0.044 & 0.661 & 0.486 & 0.729 & 0.071  & 0.698 & 0.613 & 0.761 & 0.094  \\
  NLDF\cite{nldf_sal}& 0.665 & 0.564 & 0.664 & 0.123 & 0.798 & 0.714 & 0.809 & 0.063 & 0.701 & 0.539 & 0.709 & 0.059  & 0.738 & 0.657 & 0.748 & 0.083 \\
  EGNet \cite{zhao2019EGNet} & 0.737 & 0.655 & 0.758 & 0.102 & 0.856 & 0.766 & 0.883 & 0.049 & 0.751 & 0.595 & 0.793 & 0.053  & 0.796 & 0.718 & 0.830 & 0.067  \\
  SSAL\cite{zhang2020weakly} & 0.644 & 0.579 & 0.721 & 0.126 & 0.757 & 0.702 & 0.849 & 0.071 & 0.668 & 0.527 & 0.768 & 0.066  & 0.699 & 0.647 & 0.778 & 0.092   \\ 
  SINet \cite{fan2020camouflaged} & 0.745 & 0.702 & 0.804 & 0.092 & 0.872 & 0.827 & 0.936 & 0.034 & 0.776 & 0.679 & 0.864 & 0.043 & 0.810 & 0.772 & 0.873 & 0.057 \\ \hline
  Ours\_cod\_full  & \textbf{0.793} & \textbf{0.725} & \textbf{0.826} & \textbf{0.085} & \textbf{0.893} & \textbf{0.839} & \textbf{0.938} & \textbf{0.033} & \textbf{0.793} & \textbf{0.685} & \textbf{0.868} & \textbf{0.041}  & \textbf{0.839} & \textbf{0.779} & \textbf{0.883} & \textbf{0.053}  \\ 
   \hline
  \end{tabular}
  \label{tab:benchmark_model_comparison}
\end{table*}

\begin{table*}[t!]
  \centering
  \scriptsize
  \renewcommand{\arraystretch}{1.0}
  \renewcommand{\tabcolsep}{1.3mm}
  \caption{Ablation experiments of the proposed model.}
  \begin{tabular}{l|cccccccc|cccc|cc}
  \hline
&\multicolumn{8}{c|}{Metrics for FIX}
&\multicolumn{4}{c|}{Metrics for COD}
&\multicolumn{2}{c}{Metrics for Ranking}
\\
  \multicolumn{1}{l|}{Model}&$SIM\uparrow$ & $CC\uparrow$ & $EMD\downarrow$ & $KLD\downarrow$ & $NSS\uparrow$ & $AUC\_J\uparrow$ & $AUC\_B\uparrow$ & $sAUC\uparrow$&$S_{\alpha}\uparrow$ & $F^{\mathrm{mean}}_{\beta}\uparrow$ & $E^{\mathrm{mean}}_{\xi}\uparrow$ & $\mathcal{M}\downarrow$ & $MAE\downarrow$ & $r_{MAE}\downarrow$\\
  \hline
  FIX  & 0.619 &  0.765 & 3.398 & 1.457 & 2.567 &  0.892 &  0.842 & 0.644&\ddag&\ddag& \ddag & \ddag & \ddag & \ddag \\
  COD   & \ddag & \ddag & \ddag & \ddag & \ddag & \ddag & \ddag & \ddag& 0.723 & 0.542 & 0.808 & 0.052 &\ddag&\ddag \\
  Ranking  & \ddag & \ddag & \ddag & \ddag & \ddag & \ddag & \ddag & \ddag& \ddag & \ddag & \ddag & \ddag& \textbf{0.046} & 0.143 \\ 
    Ours   & \textbf{0.622} &  \textbf{0.776} & \textbf{3.398} & \textbf{0.995} & \textbf{2.608} &  \textbf{0.901} &  \textbf{0.844} & \textbf{0.658} & \textbf{0.756} & \textbf{0.594} & \textbf{0.824} & \textbf{0.045} & 0.049 & \textbf{0.139} \\
   \hline
  \end{tabular}
  \label{tab:ablation_study}
  \vspace{-3mm}
\end{table*}
\noindent\textbf{Competing methods:}
As the number of the competing methods (SINet \cite{fan2020camouflaged} is the only deep model with code and camouflage maps available) is too
limited,
and considering the similarity of salient object detection and camouflaged object detection,
we re-train state-of-the-art salient object detection models on the
camouflaged object detection dataset \cite{fan2020camouflaged}, and treat them as competing methods.
As there exist no camouflaged object ranking models, we then implement three rank or instance based object segmentation methods for camouflage rank estimation, including RSDNet \cite{amirul2018revisiting} for salient ranking prediction, SOLOv2 \cite{wang2020SOLOv2} and Mask Scoring-RCNN (MS-RCNN) \cite{liu2018path} for instance segmentation. For the discriminative region localization task, we provide baseline performance. 

\subsection{Performance comparison}

\noindent\textbf{Discriminative region localization:}
We show the discriminative region of camouflaged objects in the first row of
Fig.~\ref{fig:predictions_visualization}, which indicates that the discriminative region, \eg heads of animals and salient patterns, could be correctly identified.
Furthermore, we show the baseline performance in
Table~\ref{tab:fixation_baseline} to quantitatively evaluate our method.


\noindent\textbf{Camouflaged object detection:}
We show the camouflaged detection map in the second row of Fig.~\ref{fig:predictions_visualization}, which is trained using our ranking dataset. We further show the quantitative results in Table~\ref{tab:benchmark_model_comparison_rankingg_dataset}, where the competing methods are re-trained using our ranking dataset. Our results in both
Fig.~\ref{fig:predictions_visualization} and
Table~\ref{tab:benchmark_model_comparison_rankingg_dataset} illustrate the effectiveness of our solution.
Moreover, as the only code-available camouflaged model, \eg SINet \cite{fan2020camouflaged}, is trained with 4,040 images from COD10K \cite{fan2020camouflaged} and CAMO \cite{le2019anabranch}, for a fair comparison, we also train our camouflaged object detection branch with the 4,040 images, and show performance in Table~\ref{tab:benchmark_model_comparison}, which further illustrates effectiveness of our method.
 

\noindent\textbf{Camouflaged object ranking:}
We show the ranking prediction in the third row of Fig.~\ref{fig:predictions_visualization}. The stacked representation of the ground truth in RSDNet is designed specifically for salient objects. We rearrange the stacked masks based on the assumption that the higher degree of camouflage corresponds to the lower degree of saliency. As is shown in Table~\ref{tab:ranking_comparison}, the performance of MS-RCNN is inferior to our method in both $MAE$ and $r_{MAE}$. Besides, although SOLOv2 achieves comparable performance with ours in terms of $MAE$, its ranking performance in $r_{MAE}$ is far from satisfactory. In order to determine the saliency rank, RSDNet borrows the instance-level ground truth to compute and descend average saliency scores of instances in an image. Therefore, the ranking is unavailable if there exists no instance-level ground truth. While analysing the model setting and performance in
Table~\ref{tab:ranking_comparison},
we clear observe the superior performance of the ranking model we proposed.

\subsection{Ablation Study}
We integrate three different tasks in our framework to achieve simultaneous discriminative region localization, camouflaged object detection and camouflaged object ranking. We then train them separately on the ranking dataset to further evaluate our solution, and show the performance on our ranking testing set in Table~\ref{tab:ablation_study}. Since the experiment for each task does not have values on metrics for the other two tasks, we use $\ddag$ to denote that the value is unavailable. For the discriminative region localization model (\enquote{FIX}), we keep the backbone network with the \enquote{Fixation Decoder} in Fig.~\ref{fig:joint_cod_fixation}. For the camouflaged object detection model (\enquote{COD}), as illustrated above, we keep the backbone network with the \enquote{Camouflage Decoder}. For the ranking model, we remove the \enquote{Joint Fixation and Segmentation prediction} module in Fig.~\ref{fig:network_overview}, and train the camouflaged object ranking network alone with the ranking annotation.

In Table~\ref{tab:ablation_study}, \enquote{Ours} is achieved through jointly training the three tasks. Comparing \enquote{FIX} and \enquote{COD} with \enquote{Ours},
we observe consistently better performance of the joint fixation baseline and our joint camouflaged prediction, which explains the effectiveness of the joint learning framework. 
While, we observe similar performance of the ranking based solution alone (\enquote{Ranking} in Table~\ref{tab:ablation_study}) compared with our joint learning ranking performance (\enquote{Ours} in Table~\ref{tab:ablation_study}), which indicates that the ranking model benefits less from the other two tasks in our framework.

Further, we adopt the dual residual attention modules (DRA) \cite{fu2019dual} in our framework as shown in Fig.~\ref{fig:joint_cod_fixation} to provide both global context information and dircriminative feature representation. We then remove the DRA modules from our triple task learning framework, and observe slightly decreased performance. The main reason is that camouflaged object detection is context-based, and the effective modelling of object context information is very beneficial.




\section{Conclusion}
We introduce two new tasks for camouflaged object detection, namely camouflaged object discriminative region localization and camouflaged object ranking, along with relabeled corresponding datasets. The former aims to find the discriminative regions that make the camouflaged object detectable, while the latter tries to explain the level of camouflage. We built our network in a joint learning framework to simultaneously localize, segment and rank the camouflaged objects. Experimental results show that our proposed joint learning framework can achieve state-of-the-art performance. Furthermore, the produced discriminative region and rank map provide insights toward understanding the nature of camouflage. Moreover, our new testing dataset NC4K can better evaluate the generalization ability of the camouflaged object detection models. 


 \section*{Acknowledgements}
 \footnotesize{This research was supported in part by National Natural Science Foundation of China (61871325, 61671387, 61620106008, 61572264), National Key Research and Development Program of China (2018AAA0102803), Tianjin Natural Science Foundation (17JCJQJC43700), CSIRO's Machine Learning and Artificial Intelligence Future Science Platform (MLAI FSP). We would like to thank the anonymous reviewers for their useful feedback.}

{\small
\bibliographystyle{ieee_fullname}
\bibliography{camouflage_ref.bib}
}

\end{document}